\documentclass[acmtog, nonacm, dvipsnames]{acmart}

\setcopyright{none}

\AtBeginDocument{%
  }

\acmISBN{978-1-4503-XXXX-X/2018/06}

\acmSubmissionID{papers\_2123}

\usepackage{xargs}
\usepackage{siunitx}
\usepackage[export]{adjustbox}
\usepackage{rotating}
\usepackage{wrapfig}
\usepackage{booktabs}
\usepackage{dblfloatfix}
\usepackage{xspace}
\usepackage{placeins}
\usepackage{xcolor}
\usepackage{amsmath}
\usepackage{amsfonts}
\usepackage[normalem]{ulem}
\usepackage[ruled]{algorithm2e}
\usepackage{multirow}
\usepackage{subcaption}

\usepackage[utf8]{inputenc} % allow utf-8 input
\usepackage[T1]{fontenc}    % use 8-bit T1 fonts
\usepackage{url}            % simple URL typesetting
\usepackage{nicefrac}       % compact symbols for 1/2, etc.
\usepackage{microtype}      % microtypography
\usepackage{tabularx}
\usepackage{tabularray}
\usepackage{listings}
\usepackage[inline]{enumitem}
\usepackage{caption}
\usepackage{graphicx}
\usepackage{glossaries}
\usepackage{pgfplots}
\pgfplotsset{compat=1.18}
\usepackage{pgfplotstable}
\usepackage{filecontents}

\usepackage{mleftright}
\usepackage{derivative}
\usepackage{pifont}
\usepackage{titlesec}
\usepackage{tablefootnote}
\usepackage{soul}
\usepackage{bm}
\usepackage[bottom]{footmisc}
\usepackage{afterpage}
\usepackage{makecell}
\usepackage{hyperref}       % hyperlinks
\usepackage{enumitem}
\UseTblrLibrary{booktabs}
\UseTblrLibrary{siunitx}

\definecolor{myBlue}{HTML}{0F1A5F}
\definecolor{myRed}{HTML}{721010}
\hypersetup{
  colorlinks,
  linkcolor={myRed},
  citecolor={myBlue},
  urlcolor={blue!80!black}
  }
\glsdisablehyper
\hypersetup{breaklinks=true}
\usepgfplotslibrary{groupplots}
\usepgfplotslibrary{colormaps}
\usepgfplotslibrary{fillbetween}
\usetikzlibrary{plotmarks}
\usetikzlibrary{patterns}

\pgfplotsset{compat=1.14}	 %
\pgfplotsset{compat/show suggested version=false}
\pgfplotsset{every mark/.append style={solid}}
\mleftright

\def\commentType{1}
\ifnum\commentType=0
    \usepackage[disable]{todonotes}
    \newcommandx{\customComment}[3]{}
    \newcommandx{\customTODO}[3]{}
\fi
\ifnum\commentType=1
\newcommandx{\customComment}[3]{\textcolor{#2}{\textsl{#1: #3}}}
  \newcommandx{\customTODO}[3]{\textcolor{#2}{{\textsl{\uline{#1: #3}}}}}
\fi
\ifnum\commentType=2
  \usepackage{pdfcomment}
  \newcommandx{\customComment}[3]{\pdfcomment[icon=Comment,opacity=0.5,color=#2,author=#1]{#3}}
  \newcommandx{\customTODO}[3]{\pdfcomment[icon=Note,opacity=0.5,color=#2,author=#1]{#3}}
\fi
\ifnum\commentType=3
  \usepackage{todonotes} % Use [disable] argument to disable todonotes.
  \usepackage{silence}
  \newcommandx{\customComment}[3]{\todo[color=#2!40,size=\small]{\textbf{#1:} #3}}
  \newcommandx{\customTODO}[3]{\todo[inline,color=#2!40,size=\small]{\textbf{#1:} #3}}
\fi

\definecolor{lightcoral}{rgb}{0.94, 0.5, 0.5}

\newcommandx{\Farnood}[1]{\customComment{Farnood}{cyan}{#1}}
\newcommandx{\Morteza}[1]{\customComment{Morteza}{lightcoral}{#1}}
\newcommandx{\Tobias}[1]{\customComment{Tobias}{seagreen}{#1}}
\newcommandx{\Romann}[1]{\customComment{Romann}{magenta}{#1}}

\newcommandx{\ToDo}[1]{\customComment{ToDo}{red}{#1}}

\definecolor{C0}{rgb}{0.121569, 0.466667, 0.705882}
\definecolor{C1}{rgb}{1.000000, 0.498039, 0.054902}
\definecolor{C2}{rgb}{0.172549, 0.627451, 0.172549}
\definecolor{C3}{rgb}{0.839216, 0.152941, 0.156863}
\definecolor{C4}{rgb}{0.580392, 0.403922, 0.741176}
\definecolor{C5}{rgb}{0.549020, 0.337255, 0.294118}
\definecolor{C6}{rgb}{0.890196, 0.466667, 0.760784}
\definecolor{C7}{rgb}{0.498039, 0.498039, 0.498039}
\definecolor{C8}{rgb}{0.737255, 0.741176, 0.133333}
\definecolor{C9}{rgb}{0.090196, 0.745098, 0.811765}

\newcolumntype{Y}{>{\centering\arraybackslash}X}
\newcolumntype{C}{>{\hsize=.0\hsize\centering\arraybackslash}X}

\colorlet{LightGoldenrod}{White!40!Goldenrod}
\colorlet{LightGray}{White!90!Periwinkle}
\definecolor{LG}{gray}{0.95}

\sethlcolor{LightGoldenrod}

\definecolor{codegreen}{rgb}{0,0.6,0}
  \definecolor{codegray}{rgb}{0.5,0.5,0.5}
  \definecolor{codepurple}{rgb}{0.58,0,0.82}
  \definecolor{backcolour}{rgb}{0.95,0.95,0.92}
  \lstdefinestyle{mystyle}{
    backgroundcolor=\color{backcolour},
    commentstyle=\color{codegreen},
    keywordstyle=\color{magenta},
    numberstyle=\tiny\color{codegray},
    stringstyle=\color{codepurple},
    basicstyle=\ttfamily\footnotesize,
    breakatwhitespace=false,
    breaklines=true,
    captionpos=b,
    keepspaces=true,
    numbers=left,
    numbersep=5pt,
    showspaces=false,
    showstringspaces=false,
    showtabs=false,
    tabsize=2
  }
  
  \lstnewenvironment{mylisting}{\lstset{style=mystyle}}{}

\newcommand{\wcfg}{w}

\newcommand{\wdetail}{w_d}

\newcommand{\pred}{D_{\mtheta}(\vz_t, t, \vy)}

\newcommand{\predl}{D_c^{L}(\vz_t)}
\newcommand{\predh}{D_c^H(\vz_t)}
\newcommand{\predv}{D_c^V(\vz_t)}
\newcommand{\predd}{D_c^D(\vz_t)}

\newcommand{\preduncondl}{D_u^{L}(\vz_t)}
\newcommand{\preduncondh}{D_u^H(\vz_t)}
\newcommand{\preduncondv}{D_u^V(\vz_t)}
\newcommand{\preduncondd}{D_u^D(\vz_t)}

\newcommand{\predcfgl}{\tilde{D}_{\mathrm{CFG}}^{L}(\vz_t)}
\newcommand{\predcfgh}{\tilde{D}_{\mathrm{CFG}}^H(\vz_t)}
\newcommand{\predcfgv}{\tilde{D}_{\mathrm{CFG}}^V(\vz_t)}
\newcommand{\predcfgd}{\tilde{D}_{\mathrm{CFG}}^D(\vz_t)}
\newcommand{\predguidednew}{\tilde{D}_{\mathrm{CFG}}(\vz_t)}

\newcommand{\prednull}{D_{\mtheta}(\vz_t, t)}

\newcommand{\predcfg}{\hat{D}_{\textrm{CFG}}(\vz_t, t, \vy)}

\newcommand{\predcond}{D_c(\vz_t)}
\newcommand{\preduncond}{D_u(\vz_t)}
\newcommand{\predguided}{\hat{D}_{\textnormal{CFG}}(\vz_t)}

\DeclareMathOperator{\dwt}{\mathtt{DWT}}
\DeclareMathOperator{\idwt}{\mathtt{iDWT}}
\usepackage{amsmath,amsfonts,bm, mathtools}

\def\eqref#1{equation~\ref{#1}}
\def\1{\bm{1}}

\newcommand{\dd}{\mathrm{d}}

\def\vx{{\bm{x}}}
\def\vy{{\bm{y}}}
\def\vz{{\bm{z}}}

\def\mI{{\pmb{I}}}

\DeclareMathAlphabet{\mathsfit}{\encodingdefault}{\sfdefault}{m}{sl}
\SetMathAlphabet{\mathsfit}{bold}{\encodingdefault}{\sfdefault}{bx}{n}

\def\mepsilon{{\bm{\epsilon}}}

\def\mtheta{{\bm{\theta}}}

\newcommand{\pdata}{p_{\textnormal{data}}}
\newcommand{\trp}[1]{#1^{\top}}
\DeclarePairedDelimiterX{\infdivx}[2]{(}{)}{%
  #1\delimsize\|#2%
}

\newcommand{\ours}{HiWave\xspace} % \textsc{}
\newcommand{\pixel}{Pixelsmith\xspace} % \textsc{}

\DeclareDocumentCommand{\ex}{m o}{
   \ensuremath{ \mathbb{E}\IfValueT{#2}{_{#2}}\left[#1\right] }
}
\DeclarePairedDelimiterX\Set[1]{\lbrace}{\rbrace}%
 {  #1 }

\def\ddefloop#1{\ifx\ddefloop#1\else\ddef{#1}\expandafter\ddefloop\fi}
\def\ddef#1{\expandafter\def\csname #1bb\endcsname{\ensuremath{\mathbb{#1}}}}
\ddefloop ABCDEFGHIJKLMNOPQRSTUVWXYZ\ddefloop
\def\ddefloop#1{\ifx\ddefloop#1\else\ddef{#1}\expandafter\ddefloop\fi}
\def\ddef#1{\expandafter\def\csname #1b\endcsname{\ensuremath{\mathbf{#1}}}}
\ddefloop ABCDEFGHIJKLMNOPQRSTUVWXYZ\ddefloop
\def\ddef#1{\expandafter\def\csname #1c\endcsname{\ensuremath{\mathcal{#1}}}}
\ddefloop ABCDEFGHIJKLMNOPQRSTUVWXYZ\ddefloop
\def\ddef#1{\expandafter\def\csname #1hat\endcsname{\ensuremath{\widehat{#1}}}}
\ddefloop ABCDEFGHIJKLMNOPQRSTUVWXYZ\ddefloop
\def\ddef#1{\expandafter\def\csname hc#1\endcsname{\ensuremath{\widehat{\mathcal{#1}}}}}
\ddefloop ABCDEFGHIJKLMNOPQRSTUVWXYZ\ddefloop
\def\ddef#1{\expandafter\def\csname #1til\endcsname{\ensuremath{\widetilde{#1}}}}
\ddefloop ABCDEFGHIJKLMNOPQRSTUVWXYZ\ddefloop
\def\ddef#1{\expandafter\def\csname tc#1\endcsname{\ensuremath{\widetilde{\mathcal{#1}}}}}
\ddefloop ABCDEFGHIJKLMNOPQRSTUVWXYZ\ddefloop
\def\ddef#1{\expandafter\def\csname #1Bar\endcsname{\ensuremath{\bar{#1}}}}
\ddefloop ABCDEFGHIJKLMNOPQRSTUVWXYZ\ddefloop

\newacronym{dwt}{DWT}{discrete wavelet transform}
\newacronym{method}{HiWave}{}

\newcommand{\figIntro}{
\begin{figure}[t]
  \centering
  \begin{subfigure}{0.30\columnwidth}
    \centering
    \includegraphics[width=\linewidth]{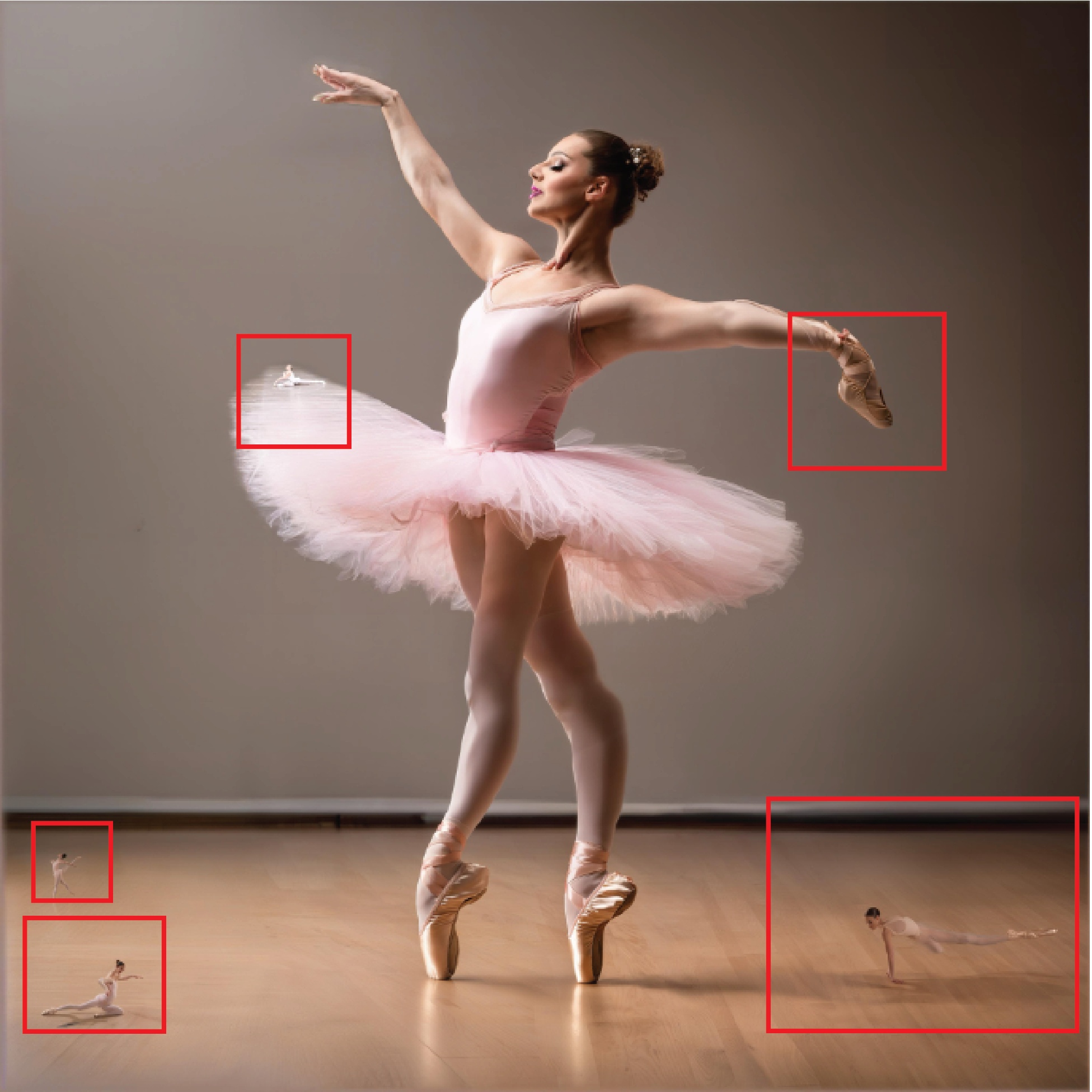}
    \subcaption{DemoFusion}
    \label{fig:demofusion}
  \end{subfigure}%
  \hfil
  \begin{subfigure}{0.30\columnwidth}
    \centering
    \includegraphics[width=\linewidth]{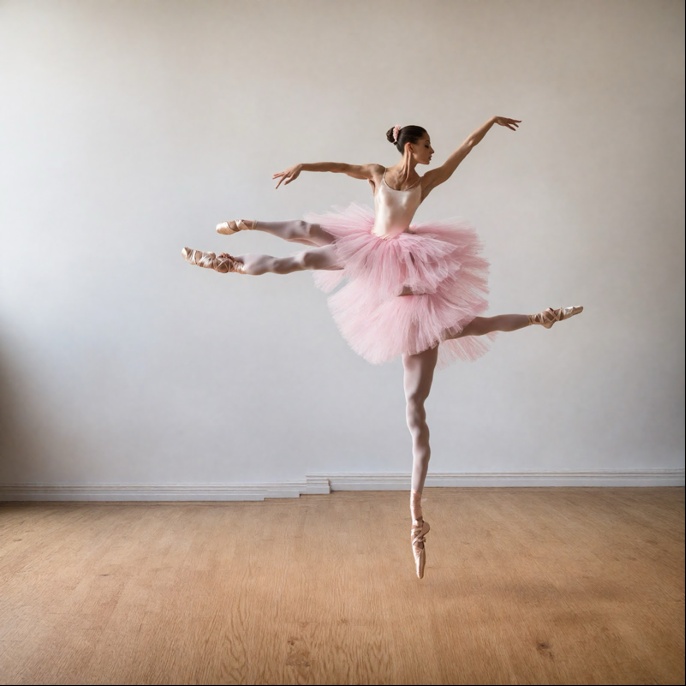}
    \subcaption{HiDiffusion}
    \label{fig:hidiffusion}
  \end{subfigure}
  \hfil
  \begin{subfigure}{0.30\columnwidth}
    \centering
    \includegraphics[width=\linewidth]{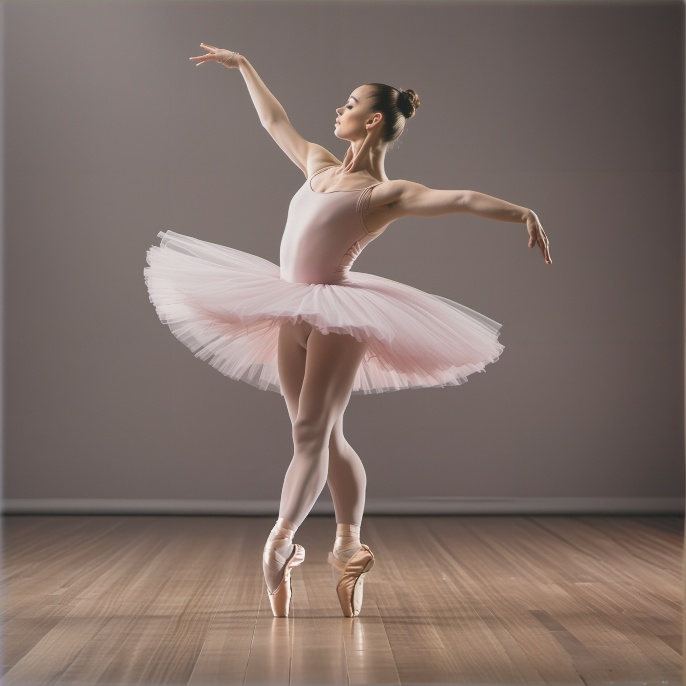}
    \subcaption{HiWave (Ours)}
    \label{fig:ourmethod}
  \end{subfigure}
  \caption{Qualitative comparison of high-resolution image generation methods at 4096$\times$4096 resolution. DemoFusion, a patch-based method, exhibits object duplication artifacts (highlighted with red rectangles). HiDiffusion, a direct inference method, lacks structural coherence and fine details. In contrast, our method produces coherent generations with rich detail.}
  \label{fig:high_res_comparison}
  \Description{}
\end{figure}
}

\newcommand{\figInterpolation}{
\begin{figure}[t]
  \centering
  \begin{subfigure}{0.48\columnwidth}
    \centering
    \includegraphics[width=\columnwidth]{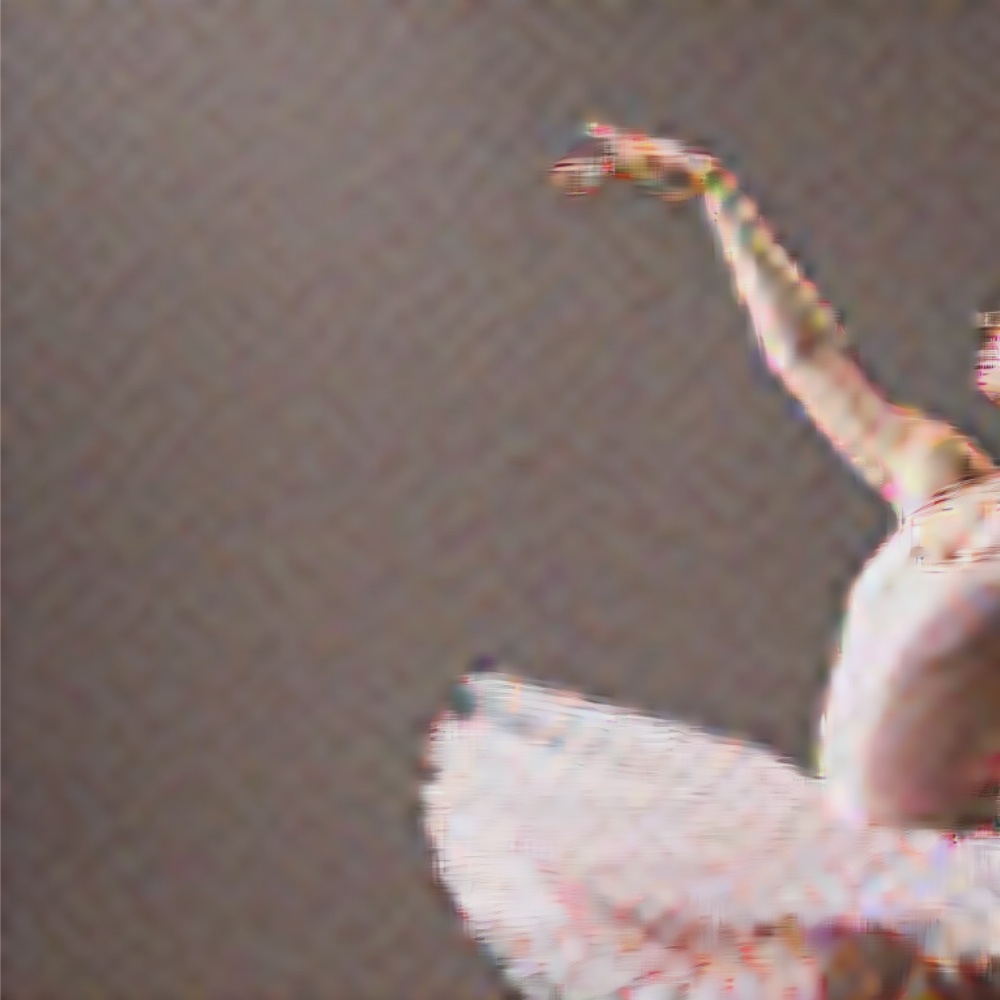}
    \subcaption{Upscaling in latent space}
    \label{fig:latent_upscaled}
  \end{subfigure}%
  \hfil
  \begin{subfigure}{0.48\columnwidth}
    \centering
    \includegraphics[width=\columnwidth]{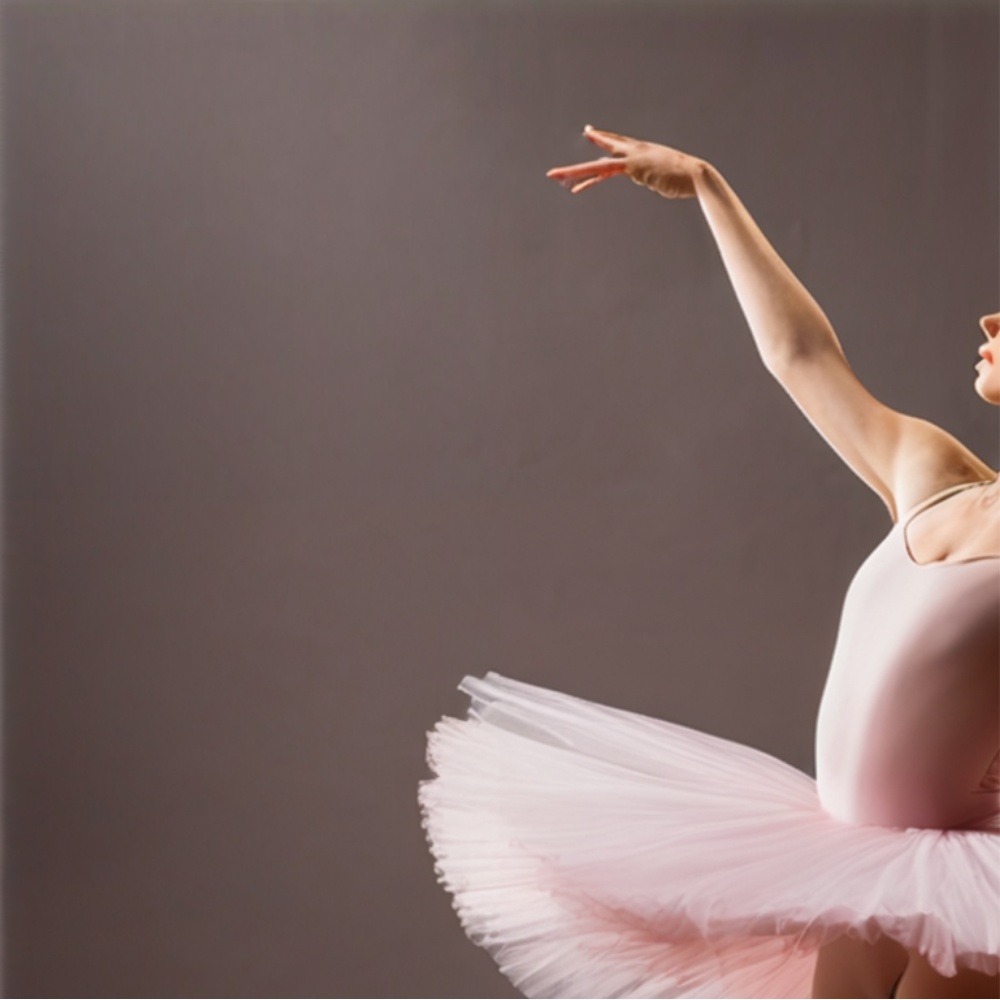}
    \subcaption{Upscaling in image space}
    \label{fig:image_upscaled}
  \end{subfigure}
  \caption{Comparison of upscaling in image space vs latent space. Interpolation performed directly in latent space introduces severe spatial artifacts, as standard VAEs are not equivariant to scaling operations. In contrast, interpolating in the image space after decoding preserves structural consistency and visual quality.}
  \label{fig:latent_vs_image_upscale}
  \Description{}
\end{figure}

}

\newcommand{\figPipeline}{
  \begin{figure*}[t]
  \centering
  \includegraphics[width=0.9\textwidth]{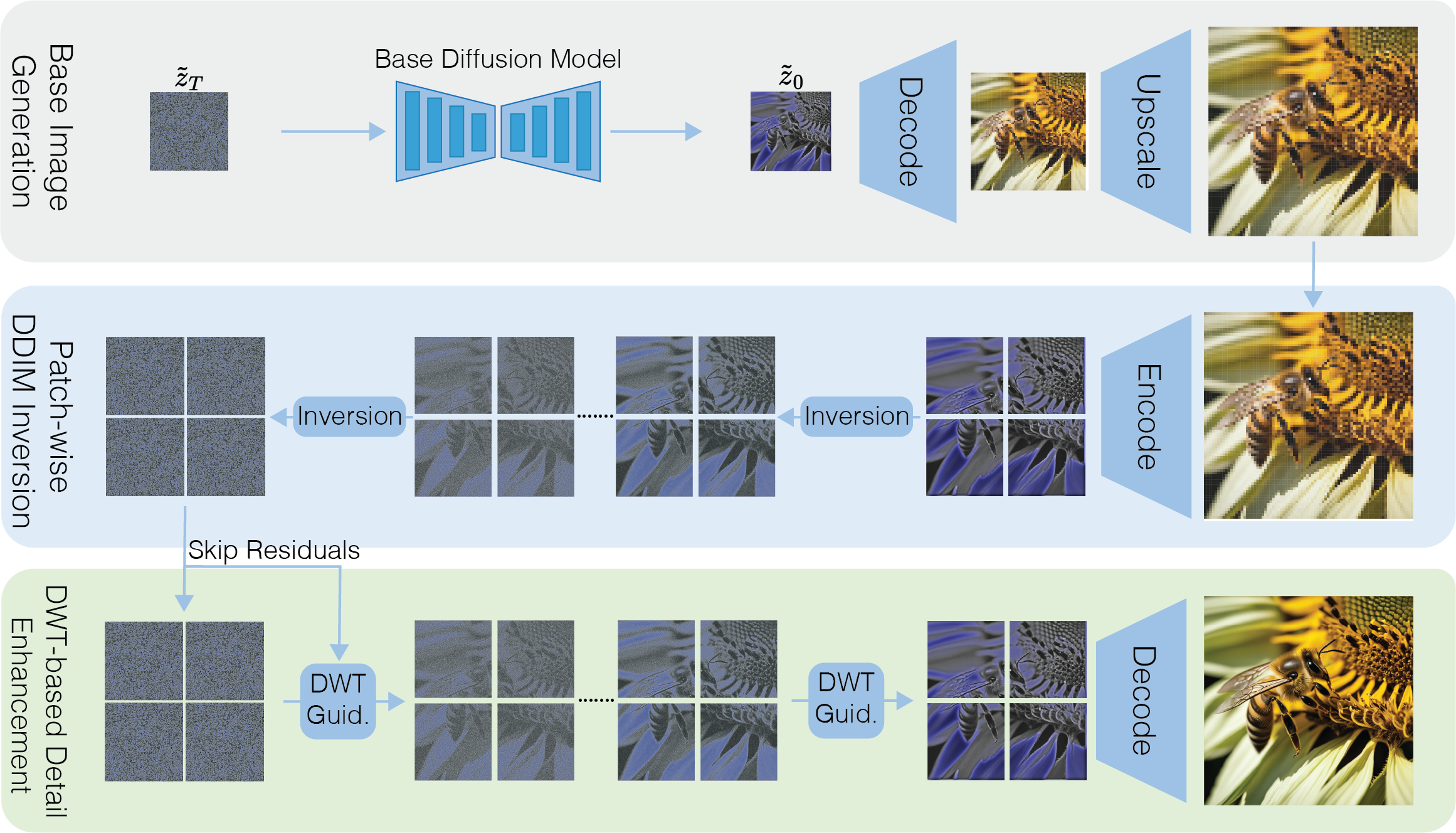}
  \caption{Overview of HiWave, our training-free high-resolution image generation pipeline. We first generate a base image using a pretrained model through a standard sampling process that transforms random noise ($z_T$) into a clean image ($\tilde{z}_0$) conditioned on a text prompt. This image is then upscaled in the image domain using Lanczos interpolation and encoded back into the latent space via the VAE encoder to enrich the base image with additional details. A patch-wise DDIM inversion process is then performed, mapping the upscaled image back to its corresponding noise representation. Finally, our DWT-based detail enhancement approach applies frequency-selective guidance during denoising, using wavelet decomposition to independently control low-frequency structure and guide high-frequency components for finer details. Skip residuals are also incorporated during the early sampling steps to further preserve the global coherence of the base image. This pipeline enables HiWave to generate high-quality, high-resolution images without duplications.}
  \label{fig:pipeline}
  \Description{}
\end{figure*}
}

\newcommand{\figMain}{
  \begin{figure*}[p]
  \centering
  \includegraphics[width=0.79\linewidth]{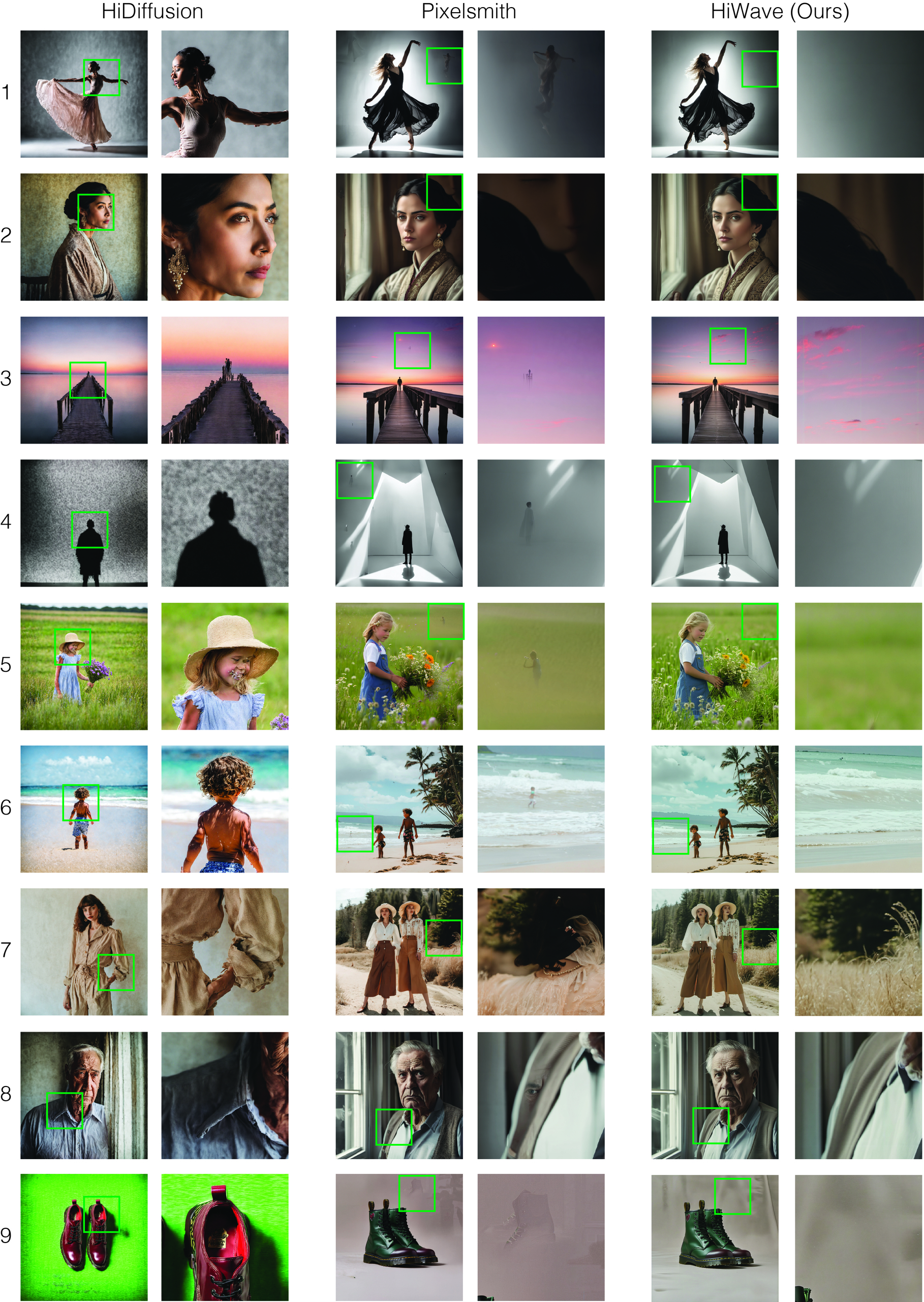}
  \caption{Qualitative comparison of high-resolution (4096$\times$4096) image generation across three methods. HiDiffusion (left column) consistently struggles to produce realistic details and coherent structures, leading to blurry textures and distorted features. Pixelsmith (middle column) generally generates high-quality details but exhibits noticeable duplication artifacts---particularly in background elements and textures---as highlighted in the zoomed regions (green boxes). In contrast, HiWave (right column) maintains structural coherence and delivers sharp, artifact-free generations without duplications.}
  \label{fig:qualitative_comparison}
  \Description{A 9$×$3 grid comparing image generation results from HiWave, Pixelsmith, and HiDiffusion at 4096$\times$4096 resolution. Each row shows a different scene with zoomed-in details. HiDiffusion consistently produces lower quality results across all examples.}
\end{figure*}
}

\newcommand{\figDetail}{
  \begin{figure*}[htp]
  \centering
  \includegraphics[width=0.95\textwidth]{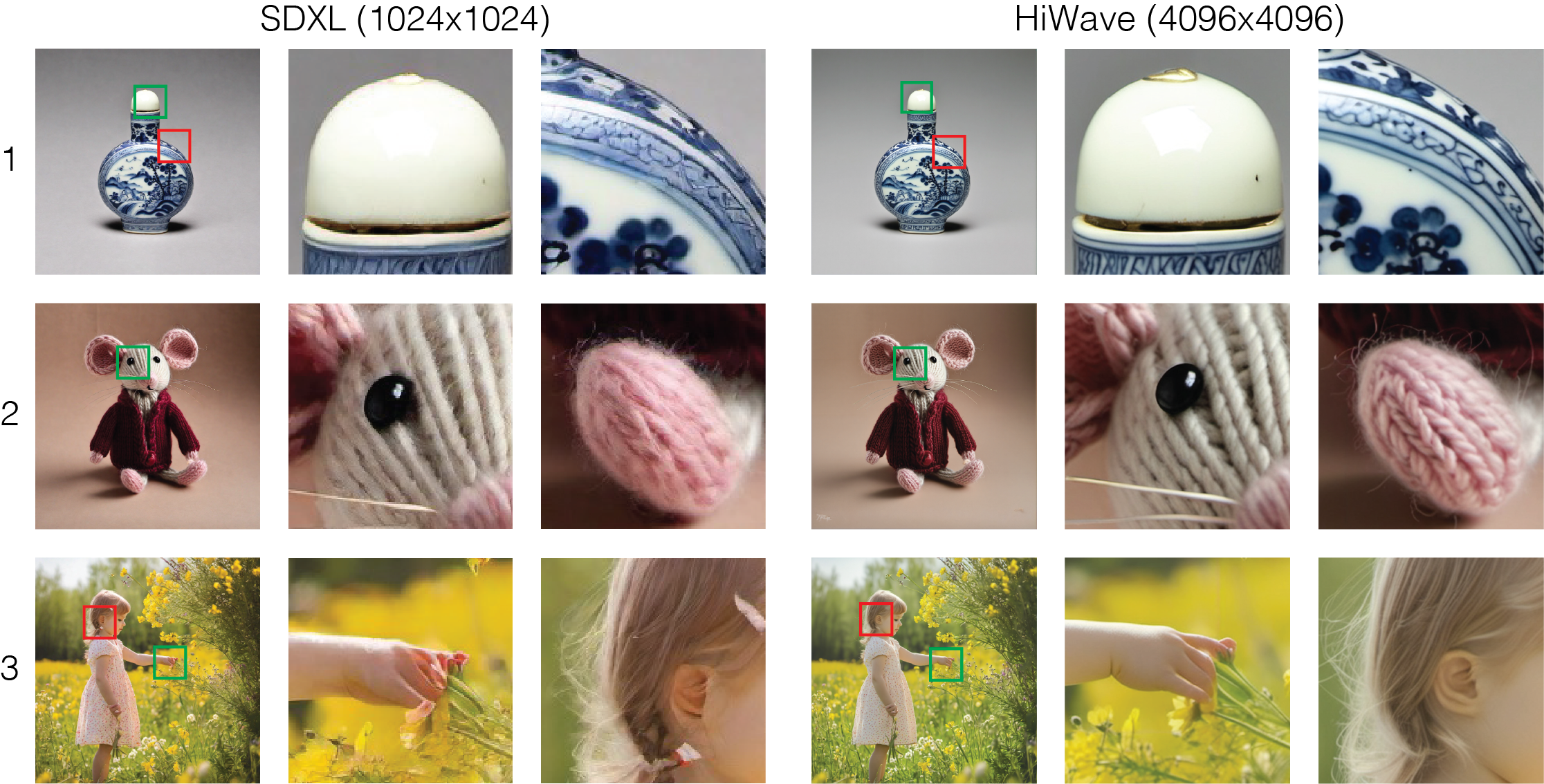}
  \caption{Comparison between our HiWave method at 4096$\times$4096 resolution (left) and the SDXL base model at 1024$\times$1024 resolution (right). Each row displays a different image along with corresponding zoomed-in regions to highlight detail enhancement. Row 1 features a porcelain bottle with intricate blue patterns; Row 2 shows a knitted toy mouse with clearly visible yarn texture; and Row 3 depicts a child in a flower field, with enhanced hair and fabric details. The zoomed regions illustrate how HiWave preserves the overall composition generated by SDXL while significantly enhancing fine details that are only partially present in the original generations.}
  \label{fig:qualitative_comparison_sdxl}
  \Description{Side-by-side comparison of images generated by HiWave at 4096$\times$4096 resolution versus SDXL at 1024$\times$1024 resolution, showing three different subjects with zoomed detail regions. The HiWave images display significantly enhanced fine details while maintaining the same overall composition as the base SDXL images.}
\end{figure*}
}

\newcommand{\figShowcase}{
  \begin{figure*}[p]
  \centering
  \includegraphics[width=0.78\textwidth]{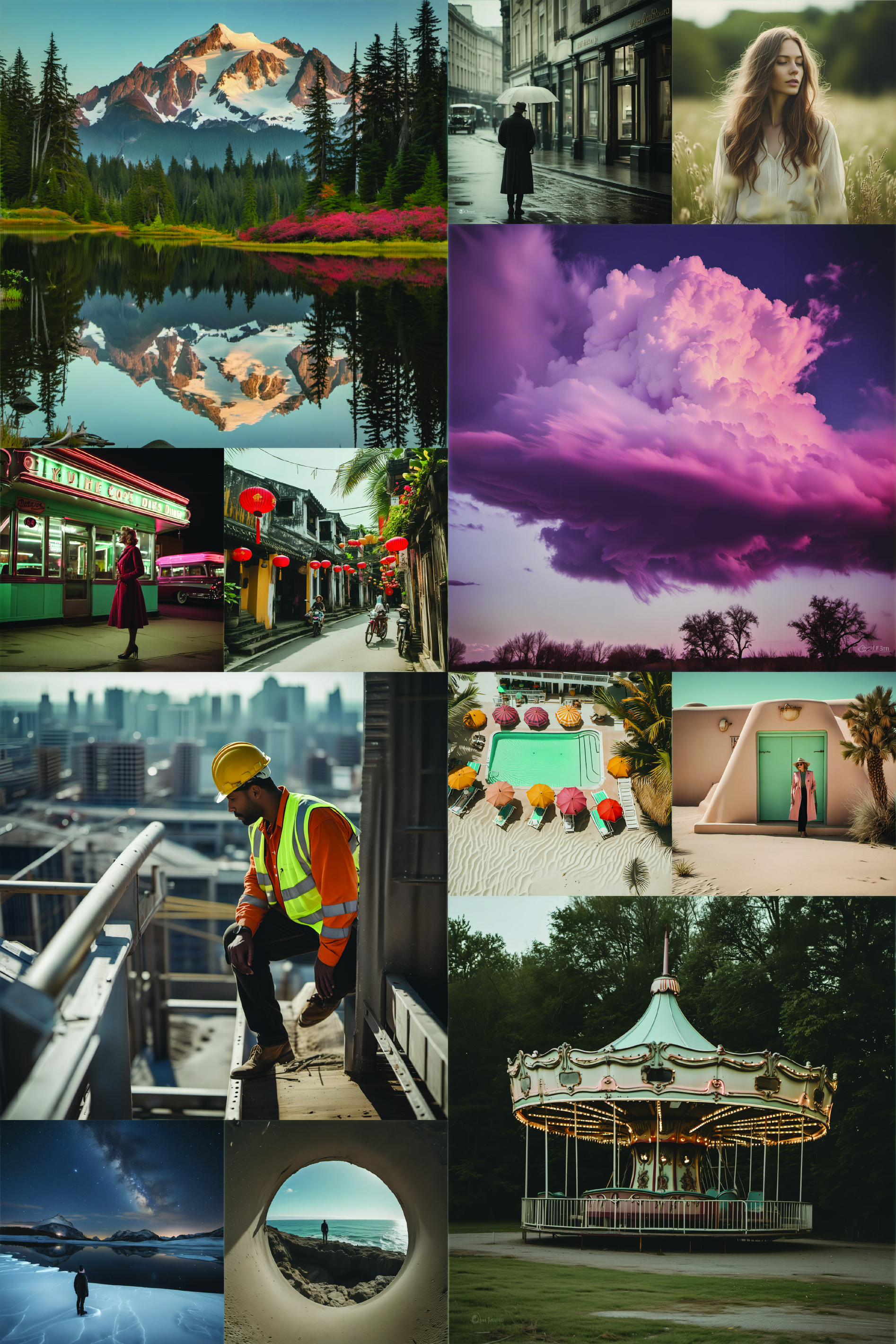}
  \caption{Examples of high-resolution (4096$\times$4096) images generated by our method, illustrating a variety of subjects across diverse visual motifs.}
  \label{fig:l}
  \Description{A collection of twelve high-resolution images demonstrating the versatility of our image generation method across diverse subjects including portraits, architectural scenes, landscapes, and artistic compositions with different lighting conditions and environments.}
\end{figure*}
}

\newcommand{\figUserStduy}{
  \begin{figure}[t]
\centering

\begin{tikzpicture}
\begin{axis}[
    xbar,
    bar width=15pt,
    xmin=0,
    xmax=100,
    xlabel={Human preference (\%)},
    symbolic y coords={HiWave, Pixelsmith},
    ytick=data,
    nodes near coords,
    nodes near coords align={horizontal},
    width=0.9\columnwidth,
    height=3.5cm,
    xtick pos=left,
    ytick pos=left,
    enlarge y limits=0.5,
    xlabel style={font=\small},
    ytick style={font=\small},
    tick label style={font=\small},
]
\addplot coordinates {(81.2,HiWave) (18.8,Pixelsmith)};
\end{axis}
\end{tikzpicture}
\vspace{-0.2cm}
\caption{User preference comparison between HiWave and Pixelsmith. HiWave was preferred by participants in over 80\% of cases, validating its effectiveness in generating high-quality images at high resolutions.}
\label{tab:barplot_preference}
\vspace{-0.35cm}
\end{figure}
}

\newcommand{\figUpscaleReal}{
  \begin{figure*}[t]
  \centering
  \includegraphics[width=0.95\textwidth]{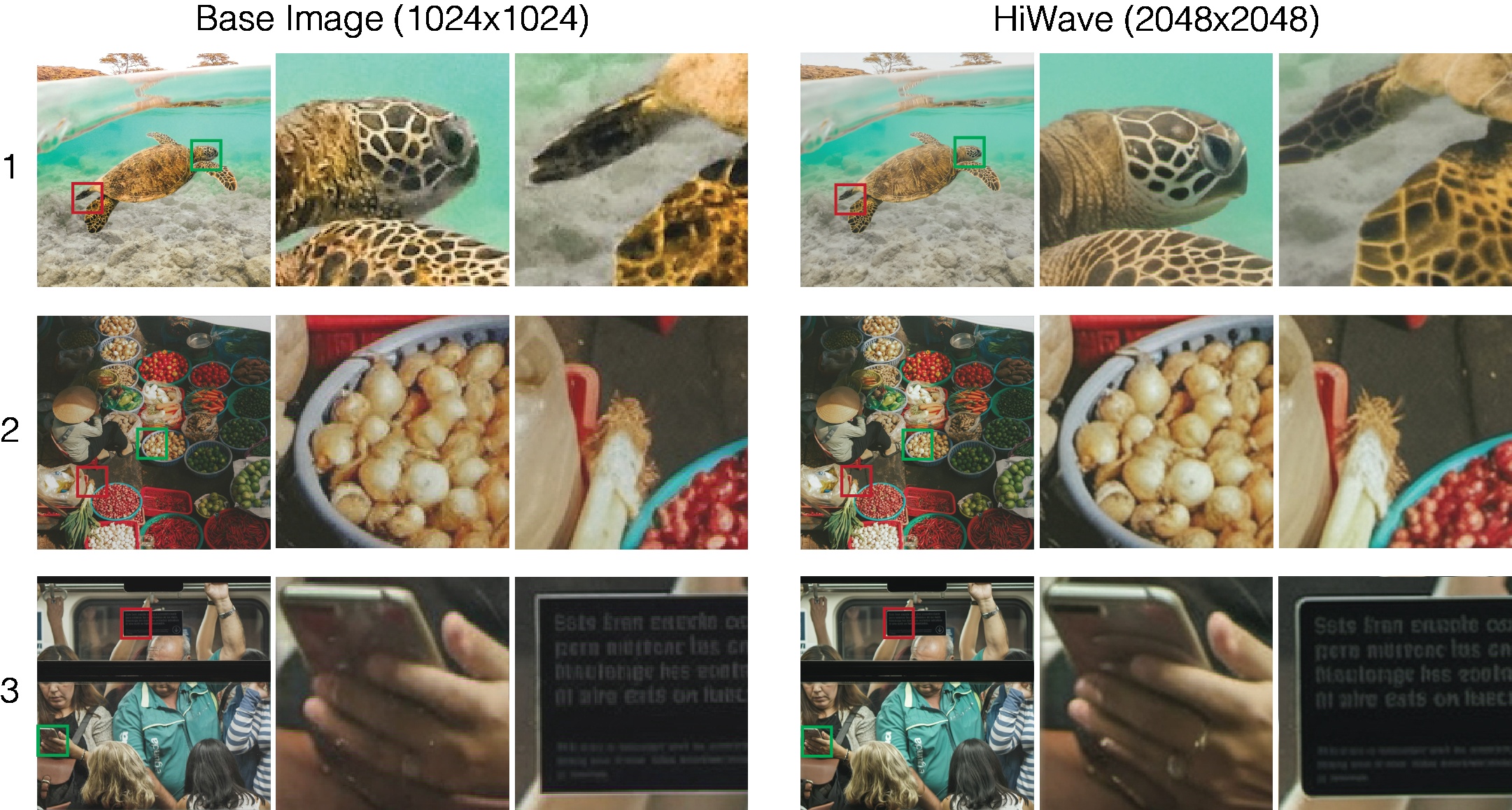}
  \caption{Upscaling natural 1024$\times$1024 photographs to 2048$\times$2048 resolution using HiWave. Each row shows one input image (left) and the corresponding HiWave result (right), along with zoomed-in crops from two marked regions (in red and green). The close-up views represent 8$\times$ magnifications. HiWave enhances fine-grained detail while preserving the semantics and appearance of the original photos.}
  \label{fig:realImagesUpscaled}
\end{figure*}
}

\begin{document}


\title{HiWave: Training-Free High-Resolution Image Generation via Wavelet-Based Diffusion Sampling}

\author{Tobias Vontobel}
\email{}
\affiliation{%
  \institution{ETH Zurich}
  \city{Zurich}
  \country{Switzerland}
}
\author{Seyedmorteza Sadat}
\email{}
\affiliation{%
  \institution{ETH Zurich}
  \city{Zurich}
  \country{Switzerland}
}
\author{Farnood Salehi}
\email{}
\affiliation{%
  \institution{Disney Research|Studios}
  \city{Zurich}
  \country{Switzerland}
}
\author{Romann Weber}
\email{}
\affiliation{%
  \institution{Disney Research|Studios}
  \city{Zurich}
  \country{Switzerland}
}

%%\renewcommand{\shortauthors}{Trovato et al.}

\begin{abstract}
Diffusion models have emerged as the leading approach for image synthesis, demonstrating exceptional photorealism and diversity. However, training diffusion models at high resolutions remains computationally prohibitive, and existing zero-shot generation techniques for synthesizing images beyond training resolutions often produce artifacts, including object duplication and spatial incoherence. In this paper, we introduce HiWave, a training-free, zero-shot approach that substantially enhances visual fidelity and structural coherence in ultra-high-resolution image synthesis using pretrained diffusion models. Our method employs a two-stage pipeline: generating a base image from the pretrained model followed by a patch-wise DDIM inversion step and a novel wavelet-based detail enhancer module. Specifically, we first utilize inversion methods to derive initial noise vectors that preserve global coherence from the base image. Subsequently, during sampling, our wavelet-domain detail enhancer retains low-frequency components from the base image to ensure structural consistency, while selectively guiding high-frequency components to enrich fine details and textures. Extensive evaluations using Stable Diffusion XL demonstrate that HiWave effectively mitigates common visual artifacts seen in prior methods, achieving superior perceptual quality. A user study confirmed HiWave's performance, where it was preferred over the state-of-the-art alternative in more than 80\% of comparisons, highlighting its effectiveness for high-quality, ultra-high-resolution image synthesis without requiring retraining or architectural modifications.
\end{abstract}

% \keywords{Image synthesis, diffusion models, high-resolution generation, discrete wavelet transform}

\begin{teaserfigure}
  \centering
  \includegraphics[width=\textwidth]{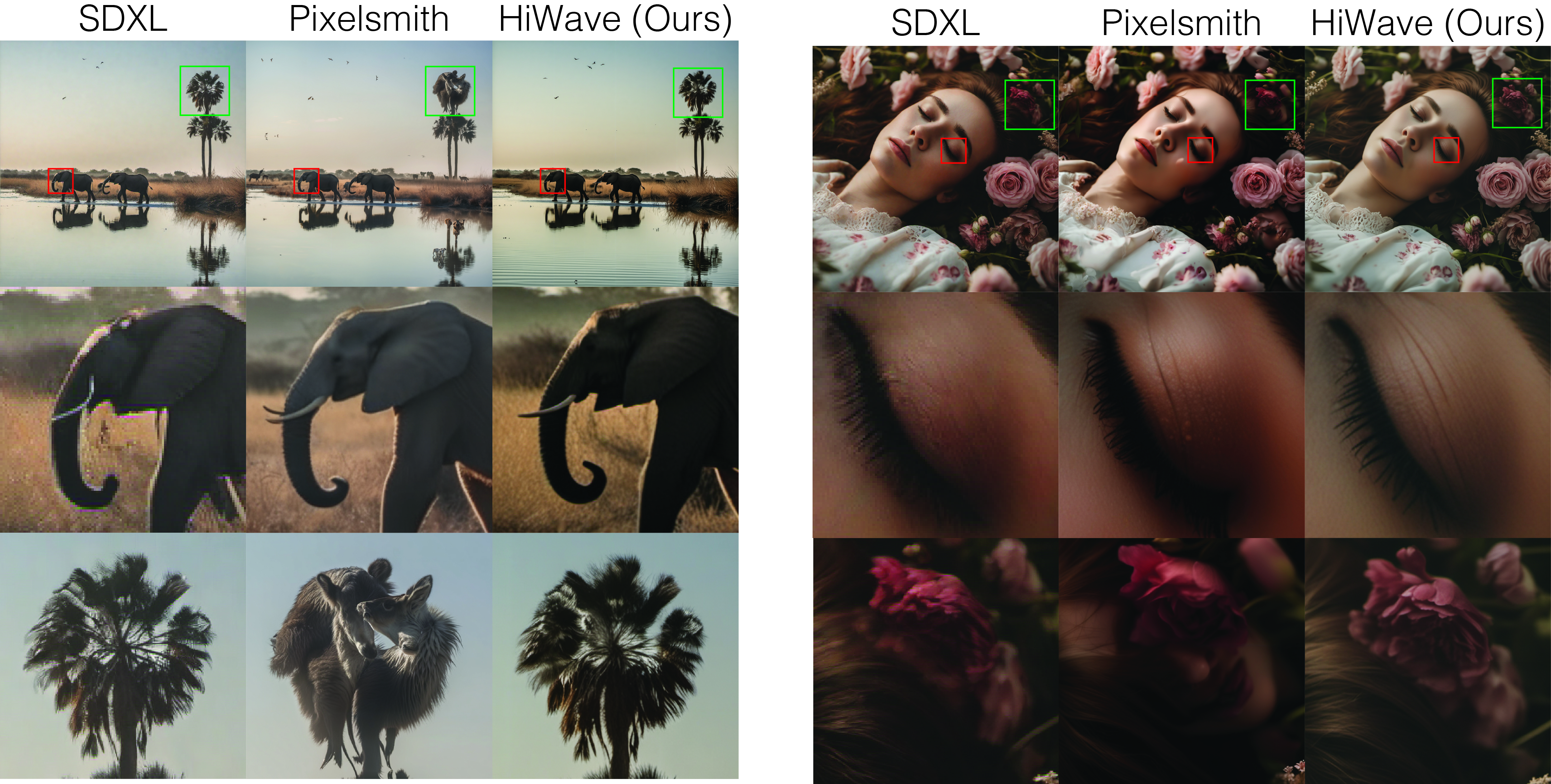}
  \caption{We propose \gls{method}, a novel training-free approach for high-resolution image generation using pretrained diffusion models. While standard Stable Diffusion XL (SDXL) can produce globally coherent images, it lacks fine details when upscaled to 4096$\times$4096 resolution (left column). Existing training-free methods  (e.g., Pixelsmith \citep{tragakis2024one}) enhance details in SDXL outputs but often introduce duplicated objects and visual artifacts (middle column). In contrast, \gls{method} leverages a patch-wise DDIM inversion strategy combined with a wavelet-based detail enhancer module to produce high-quality images with rich details and minimal duplication artifacts. The second and third rows show 10$\times$ and 5$\times$ magnified views of the red and green boxed regions, respectively.}
  \label{fig:teaser}
  \vspace{0.5cm}
\end{teaserfigure}



\maketitle

\section{Introduction}
\label{sec:introduction}

Since the introduction of diffusion models, generative image synthesis has reached unprecedented levels of photorealism and creative control. Recent models such as Stable Diffusion \cite{podell2023sdxl, esser2024scaling} can generate stunning images at resolutions up to 1024$\times$1024 pixels. Despite these improvements in image quality, producing outputs beyond 1024$\times$1024 remains technically challenging due to the substantial computational demands associated with training at higher resolutions.

Diffusion models typically rely on large-scale networks with high parameter counts to achieve optimal image quality and prompt alignment \cite{esser2024scaling}. As input resolution increases, the computational cost of training these models becomes prohibitive---especially for attention-based architectures \cite{peebles2023scalable,esser2024scaling}, where complexity scales quadratically with spatial dimensions. Moreover, most datasets used to train large-scale diffusion models lack high-resolution content beyond 1024$\times$1024. As a result, current state-of-the-art models are typically limited to moderate resolutions, restricting their applicability in domains such as advertising and film production, where ultra high-resolution outputs (e.g., 4K) are required.

Motivated by these limitations, recent work has explored methods to extend pretrained diffusion models to higher resolutions---i.e., beyond their native training sizes. These approaches fall into two main categories: \emph{patch-based} techniques that process image regions independently, such as Pixelsmith \cite{tragakis2024one} and DemoFusion \cite{du2024demofusion}, and \emph{direct inference} methods that modify model architectures, such as HiDiffusion \cite{zhang2023hidiffusion} and FouriScale \cite{huang2024fouriscale}. However, these methods face fundamental limitations. Patch-based approaches often produce duplicated objects (\autoref{fig:demofusion}), while direct inference methods struggle to maintain global coherence at very high resolutions (e.g., beyond 2048$\times$2048, see \autoref{fig:hidiffusion}). This highlights the need for a training-free, high-resolution generation approach that ensures both global coherence and robustness to duplication and artifacts.

In this paper, we propose HiWave, a novel pipeline for training-free high-resolution image generation that produces globally coherent outputs without object duplication. HiWave adopts a two-stage, patch-based approach that preserves the coherent structure of a base image generated by a pretrained model, while enhancing the fine details required for higher resolutions. In the first stage, a base image is generated at a standard resolution (e.g., 1024$\times$1024) using a pretrained diffusion model. This image is then upscaled in the image domain to the target high resolution, though the upscaled result lacks fine-grained details. To enrich the image with high-frequency details, we introduce a novel sampling module based on patch-wise DDIM inversion. Specifically, each image patch is inverted using DDIM to recover the corresponding latent noise that would generate the given input. Sampling is then performed starting from this inverted noise. However, to prevent the model from simply reproducing the original image, we incorporate a \emph{detail enhancer} into the sampling process. This component is based on \gls{dwt}, and it preserves the low-frequency content of the base image to maintain global structure while guiding the high-frequency components to synthesize realistic additional details suitable for the target resolution.

Our approach enables standard diffusion models trained at a resolution of 1024$\times$1024 to generate images at 4096$\times$4096 resolution---a 16$\times$ increase in total pixel count. HiWave leverages the high-frequency priors inherently captured by pretrained diffusion models, as observed by \citet{du2024demofusion}, to generate coherent images with fine details. Compared to existing methods, HiWave excels at maintaining structural coherence while significantly reducing hallucinations and duplicated artifacts, addressing a key limitation in current zero-shot high-resolution generation pipelines. We validate the effectiveness of HiWave using Stable Diffusion XL \cite{sdxl} and compare it against \pixel \cite{tragakis2024one}, the current state-of-the-art in training-free high-resolution image generation. In a user study, HiWave was preferred over \pixel in more than 80\% of cases, highlighting its superior visual fidelity.

In summary, our main contributions are the following:
\begin{itemize}
  \item We introduce HiWave, a training-free, two-stage pipeline for high-resolution image synthesis that extends pretrained diffusion models to ultra high-resolutions, without requiring architectural modifications or additional training.
  \item  We propose a novel patch-wise DDIM inversion framework coupled with a wavelet-based detail enhancer, which preserves global structure from the base image while selectively enriching its high-frequency details.
  \item We demonstrate that HiWave effectively mitigates common artifacts such as duplicated objects and structural inconsistencies that persist in prior methods.
  \item  We conduct extensive evaluations with Stable Diffusion XL and show that HiWave outperforms current state-of-the-art methods in qualitative comparisons and a preference study, with users favoring HiWave outputs in over 80\% of cases.
\end{itemize}

\figIntro

\section{Related Work}
\label{sec:related}

Diffusion models have become a dominant framework for image synthesis due to their strong generative capabilities \cite{ho2020denoising,song2020score,denton2015deep}. They have rapidly surpassed previous generative modeling techniques in terms of fidelity and diversity \citep{nichol2021improved,dhariwalDiffusionModelsBeat2021}, achieving state-of-the-art performance across a wide range of applications, including unconditional image synthesis \citep{dhariwalDiffusionModelsBeat2021,karras2022elucidating}, text-to-image generation \citep{dalle2,saharia2022photorealistic,balaji2022ediffi,rombachHighResolutionImageSynthesis2022,sdxl,yu2022scaling,esser2024scaling}, video synthesis \citep{blattmann2023align,stableVideoDiffusion,gupta2023photorealistic}, image-to-image translation \citep{saharia2022palette,liu20232i2sb}, motion synthesis \citep{tevet2022human,Tseng_2023_CVPR}, and audio synthesis \citep{WaveGrad,DiffWave,huang2023noise2music}.

Despite this progress, diffusion models often incur high computational costs and long training times \citep{chen2023pixart}, especially when handling high-resolution data. Latent diffusion models (LDMs) \citep{rombachHighResolutionImageSynthesis2022} alleviate some of this burden by compressing inputs into a smaller latent space using a pretrained autoencoder. However, LDMs typically scale only up to moderate resolutions (e.g., 1024$\times$1024). While some recent works have explored training LDMs at higher resolutions \citep{Chen2024PIXARTFA,xie2024sana}, they continue to face challenges related to prolonged training times and the limited availability of high-quality ultra-high-resolution datasets. These constraints have spurred growing interest in training-free methods that exploit pretrained models to generate images beyond their native resolutions.

One popular direction decomposes high-resolution generation into patches that are processed independently and later combined to form the full image. Examples include DemoFusion \cite{du2024demofusion}, AccDiffusion \cite{lin2024accdiffusion}, and Pixelsmith \cite{tragakis2024one}. This patch-based strategy allows diffusion models to operate at their native resolution for each patch, thereby preserving the detail and expressiveness of the base model. However, these methods often fail to maintain global coherence at resolutions beyond 2048$\times$2048, frequently producing boundary artifacts, duplicated content, and semantic inconsistencies across patches.

Another line of work modifies the architecture or inference process of pretrained diffusion models---without additional training---to enable single-pass generation of high-resolution images. Notable examples include FouriScale \cite{huang2024fouriscale}, MegaFusion \cite{wu2025megafusion}, and HiDiffusion \cite{zhang2023hidiffusion}. These methods introduce attention scaling or downsampling blocks into the pretrained network to better align intermediate features with those seen at the model’s native resolution. While such methods avoid patch-based artifacts, their performance often degrades at ultra-high resolutions (e.g., 4096$\times$4096). As the gap between training and inference resolutions widens, these approaches struggle to preserve global coherence---leaving patch-based strategies as the current state-of-the-art for zero-shot high-resolution image generation.

In summary, current training-free approaches either fail to maintain global coherence compared to the base diffusion model or suffer from duplicated objects and artifacts  at ultra-high resolutions (e.g., 4096$\times$4096). \gls{method} proposes a novel patch-based strategy aimed at achieving coherent, high-quality image generation at such resolutions while avoiding the limitations of existing methods.

\section{Background}
\figPipeline

\paragraph{Diffusion models}
Let $\vx \sim \pdata(\vx)$ represent a data sample, and let $t \in [0, T]$ denote a continuous time variable. In the forward diffusion process, noise is incrementally added as $\vz_t = \vx + \sigma(t)\mepsilon$, where $\sigma(t)$ is a time-dependent noise scale. This noise schedule gradually corrupts the data, with $\sigma(0) = 0$ (clean data) and $\sigma(T) = \sigma_{\textnormal{max}}$ (maximum corruption). As shown in \citet{karras2022elucidating}, this process corresponds to the following differential equation:
\begin{equation}\label{eq:diffusion-ode}
\dd\vz = - \dot{\sigma}(t) \sigma(t) \nabla_{\vz_t} \log p_t(\vz_t) \dd t,
\end{equation}
where $p_t(\vz_t)$ denotes the distribution over noisy samples at time $t$, transitioning from $p_0 = \pdata$ to $p_T = \mathcal{N}(\mathbf{0}, \sigma_{\textnormal{max}}^2 \mI)$. Sampling from the data distribution involves solving this ODE in reverse (from $t = T$ to $t = 0$), provided the score function $\nabla_{\vz_t} \log p_t(\vz_t)$ is known. Since this score function is intractable, it is approximated by a neural denoiser $D_{\mtheta}(\vz_t, t)$, trained to reconstruct the clean signal $\vx$ from its noisy observation $\vz_t$. For conditional generation, the denoiser is extended with a conditioning variable $\vy$---such as a label or text prompt---resulting in $D_{\mtheta}(\vz_t, t, \vy)$.

\paragraph{Inversion methods}
Inversion techniques, such as DDIM inversion \citep{songDenoisingDiffusionImplicit2022} solve the forward-time version of the diffusion ODE in \autoref{eq:diffusion-ode} to map an observed image $\vx$ to its corresponding noise vector $\vz_T$. By integrating \autoref{eq:diffusion-ode} from $t = 0$ to $t = T$, one obtains a deterministic mapping from the image back to the noise space (in the limit of small steps). This enables applications like image editing, as it provides a way to anchor sampling around a given input while retaining its global layout.

\paragraph{Classifier-free guidance}
Classifier-free guidance (CFG) \citep{hoClassifierFreeDiffusionGuidance2022} is an inference-time technique that improves generation quality by blending predictions from conditional and unconditional models. During sampling, CFG adjusts the denoiser output as:
\begin{equation}\label{eq:cfg}
\predcfg = \prednull + \wcfg (\pred - \prednull),
\end{equation}
where $\wcfg$ is a guidance strength parameter (with $\wcfg = 1$ representing unguided sampling). The unconditional model $\prednull$ is typically learned by randomly dropping conditioning inputs $\vy$ during training. Alternatively, separate unconditional models can be used \citep{karras2023analyzing}. Analogous to the truncation trick in GANs \citep{brockLargeScaleGAN2019}, CFG improves visual fidelity, but may lead to oversaturation \citep{sadat2025eliminating} or reduced diversity \citep{sadat2024cads}.

\paragraph{Discrete wavelet transform}
Discrete wavelet transforms (DWT) \citep{waveletIntro} are a fundamental tool in signal processing, commonly used to analyze spatial-frequency content in data. The transformation utilizes a pair of filters---a low-pass filter $L$ and a high-pass filter $H$. For 2D signals, these are combined to form four distinct filter operations: $L\trp{L}$, $L\trp{H}$, $H\trp{L}$, and $H\trp{H}$. When applied to an image $\vx$, the 2D wavelet transform decomposes it into one low-frequency component $\vx_{L}$ and three high-frequency components ${\vx_{H}, \vx_{V}, \vx_{D}}$, capturing horizontal, vertical, and diagonal details, respectively. Each of these sub-bands has spatial dimensions of $H/2 \times W/2$ for an image of size $H \times W$. Multiscale decomposition can be achieved by recursively applying the transform to the low-frequency component $\vx_{L}$. The transformation is fully invertible, enabling exact reconstruction of the original image $\vx$ from the set $\Set{\vx_{L}, \vx_{H}, \vx_{V}, \vx_{D}}$ using the inverse discrete wavelet transform (iDWT). 
% Additionally, fast wavelet transform (FWT) \citep{mallat1989theory} allows efficient computation of these sub-bands in linear time with respect to the number of pixels in $\vx$.

\section{Method}
\label{sec:method}
We now describe the details of HiWave, our framework for high-resolution image generation using pretrained diffusion models. An overview of the complete pipeline is shown in \autoref{fig:pipeline}, and we detail each component below. We define $\predcond \doteq \pred$, $\preduncond \doteq \prednull$, and $\predguided \doteq \predcfg$ to represent the conditional, unconditional, and CFG outputs, respectively.

\subsection{Base image generation}
\figInterpolation
We begin by generating a base image at the native resolution of the pretrained diffusion model, typically 1024$\times$1024. This image is then upscaled in the image domain using Lanczos interpolation. Unlike most previous works that perform upscaling in the latent space \citep{du2024demofusion}, we opt for image-domain upscaling to avoid artifacts commonly introduced by latent-space interpolation. These artifacts arise because standard VAEs used in diffusion pipelines are not equivariant to scaling operations, leading to inconsistencies when upscaling is applied in latent space \cite{kouzelis2025eq}. An example of such artifacts is illustrated in \autoref{fig:latent_vs_image_upscale}. To avoid this, we first upscale the base image to the target resolution (e.g., 4096$\times$4096) in the image domain.
At this point, we obtain an image at the target resolution, albeit lacking fine-grained details. We then encode this upscaled image into the latent space via the VAE encoder and apply a patch-based sampling process (described next) to refine high-resolution details while preserving the global structure of the original image.

\subsection{Patch-wise DDIM inversion}
To preserve structural coherence during patch-wise generation, we initialize the diffusion process using DDIM inversion \citep{song2020denoising} instead of random noise. This inversion retrieves noise vectors for each image patch by integrating the diffusion ODE forward in time:
\begin{equation}
\vz_{t+1} \approx \vz_t - \dot{\sigma}(t) \sigma(t) \nabla_{\vz_t} \log p_t(\vz_t) \Delta t,
\end{equation}
This deterministic initialization provides two key benefits:
\begin{enumerate*}
\item {Controlled noise for frequency decomposition:} DDIM-inverted noise retains meaningful structural information and spatial layout from the original image, enabling our detail enhancement module to selectively refine high-frequency textures while preserving the low-frequency structural content.
\item {Consistent patch initialization:} Neighboring patches receive compatible noise vectors, which helps maintain continuity and avoid visible seams across patch boundaries.
\end{enumerate*}
In contrast, initializing with random Gaussian noise loses the structural context of the base image and often introduces artifacts or structural inconsistencies. This controlled inversion lays the groundwork for our DWT-based detail enhancer in the next step, enabling detail enhancement while preserving global coherence.

\subsection{Detail enhancement with DWT guidance}
In the next step, we begin sampling from the inverted noise of each patch, with the goal of progressively adding details to the final image. A central challenge in patch-based high-resolution generation is maintaining a balance between global coherence and detailed texture synthesis. Existing methods often fall short in one of these areas, resulting in either artifact-prone high-frequency patterns or globally consistent images that lack detail.

To address this, we introduce a DWT-based detail enhancement module that leverages the complementary roles of the low- and high-frequency components of each latent. We argue that low-frequency components typically capture structural coherence, while high-frequency components convey fine details and textures. This distinction is especially critical in patch-wise generation, where each patch must integrate seamlessly into the global image for overall coherence while still containing sufficient detail at high resolutions.

Since we used DDIM inversion to obtain the final noise, following the sampling process using only the conditional prediction $\predcond$ would reproduce the base image, i.e., globally coherent but lacking in fine detail. This motivates our decision to preserve the low-frequency bands from $\predcond$, which retain much of the base image’s global layout due to the DDIM inversion. To enrich each patch with the required details for high-quality generation, we guide the high-frequency components adaptively using a modified CFG strategy.

Specifically, we apply the \gls{dwt} to both the conditional and unconditional predictions to get
\begin{align}
\dwt(\predcond) &= \Set{\predl, \predh, \predv, \predd}, \\
\dwt(\preduncond) &= \Set{\preduncondl, \preduncondh, \preduncondv, \preduncondd}.
\end{align}
We then construct the guided prediction in the frequency domain as follows:
\begin{align}
\predcfgl &= \predl, \\
\predcfgh &= \preduncondh + \wdetail (\predh - \preduncondh), \\
\predcfgv &= \preduncondv + \wdetail (\predv - \preduncondv), \\
\predcfgd &= \preduncondd + \wdetail (\predd - \preduncondd).
\end{align}
Finally, we apply the inverse DWT to reconstruct the full guided signal:
\begin{equation}
\predguidednew = \idwt(\Set{\predcfgl, \predcfgh, \predcfgv, \predcfgd}).
\end{equation}
This frequency-aware guidance strategy enables precise enhancement of details while preserving the global structure of the base image.

\subsection{Skip residuals}
To further preserve global structure during early denoising, we incorporate skip residuals by mixing the latents obtained from DDIM inversion with those from the sampling process for each patch. Let $\vz_t$ denote the current latent in the sampling process, $\vz^s_t$ the corresponding DDIM-inverted latent, $\tau$ a time step threshold, and $c_1 = ((1 + \cos(\frac{T-t}{T} \pi))/2)^\alpha$ a cosine-decay weighting factor. The skip residual update is defined as
\begin{equation}
\hat{\vz}_t = \begin{cases}
c_1 \times \vz_t + (1 - c_1) \times \vz^s_t & \quad t < \tau, \\
\vz_t & \quad t \geq \tau.
\end{cases}
\end{equation}
Unlike prior work that applies skip residuals throughout all diffusion steps, we adopt a more conservative strategy: they are only used during the initial denoising phase. This allows the model to leverage the base image’s structure early on, and then progressively diverge to synthesize novel details guided by the DWT-enhanced predictions. 
In contrast, applying skip residuals at all time steps--as done in previous work--can suppress detail synthesis, while omitting them entirely causes duplication artifacts.
Our DWT-based enhancer mitigates this trade-off by explicitly guiding different frequency bands, enabling the generation of rich textures while preserving global consistency.

\subsection{Implementation Details}
We use the \texttt{sym4} wavelet for DWT due to its effective balance between spatial and frequency localization. The detail guidance strength is set to $\wdetail = 7.5$, enhancing high-frequency features while preserving the low-frequency structure from $\predcond$.

Image generation is performed progressively---first at 1024$\times$1024 (the model’s native resolution), then at 2048$\times$2048, and finally at 4096$\times$4096. While some prior works report increased duplication artifacts with iterative upscaling \cite{tragakis2024one}, we did not observe this in our experiments, likely due to the combination of patch-wise DDIM inversion and our DWT-based guidance.

Skip residuals are applied only up to time step 15 (out of 50) for 2048$\times$2048 resolution and up to time step 30 for 4096$\times$4096. This conservative strategy contrasts with methods that apply skip residuals across the entire diffusion process. By limiting their use to early steps, we preserve the global structure initially while allowing the model to synthesize novel details in later time steps.

For patch processing, we use a 50\% overlap between adjacent patches to ensure smooth transitions. To optimize memory usage, we employ a streaming approach in which patches are processed in batches rather than all at once, enabling 4096$\times$4096 image generation on consumer GPUs with 24GB of VRAM.

\section{Results}
\label{sec:result}

We now evaluate \gls{method} both qualitatively and quantitatively against state-of-the-art high-resolution image generation methods. Our evaluation focuses on three main aspects:
\begin{enumerate*}
\item avoiding the duplication artifacts commonly seen in previous patch-based methods,
\item achieving global image coherence and high fidelity at 4096$\times$4096 resolution, and
\item enhancing both quality and details over the original Stable Diffusion XL outputs.
\end{enumerate*}

\subsection{Experimental setup}
We select Pixelsmith \citep{tragakis2024one} as the leading patch-based approach and HiDiffusion \citep{zhang2023hidiffusion} as the representative of direct inference methods. All methods were used to generate 4096$\times$4096 resolution images from the same prompts and identical random seeds, using Stable Diffusion XL as the base model. To ensure a fair comparison, we employed the official codebases of all baselines. All experiments were conducted on a single RTX 4090 GPU with 24GB of VRAM. 

To benchmark the methods, we used 1000 randomly sampled prompts from the LAION/LAION2B-en-aesthetic \citep{schuhmann2022laionb} dataset, covering a diverse range of content including natural landscapes, human portraits, animals, architectural scenes, and close-up textures. This diversity allowed us to assess performance across a broad set of generation challenges.

For quantitative evaluation, prior work has shown that existing metrics are often unreliable at high resolutions, as they typically downscale images to lower resolutions (e.g., 224$\times$224) before computing the score \citep{zhang2023hidiffusion,du2024demofusion,tragakis2024one}. Consequently, we rely primarily on a human study, which provides the most reliable assessment of perceptual quality in this setting. Standard quantitative metrics are also reported in \autoref{sec:quantitative-evaluation} for completeness.

\subsection{Qualitative comparison with prior methods}
\autoref{fig:qualitative_comparison} presents a comprehensive visual comparison of our method against Pixelsmith and HiDiffusion across nine diverse test cases. Each row corresponds to a different subject, with the left columns displaying the full-resolution outputs and the right columns showing magnified regions (highlighted by green boxes) to facilitate detailed inspection of fine structures. HiDiffusion produces outputs that lack coherent structure and exhibit blurred textures across all nine examples. This demonstrates that, while direct inference methods avoid duplication and patch-based artifacts, they often struggle to produce globally coherent outputs with fine-grained detail at high resolutions. In contrast, Pixelsmith generates more detailed images but frequently suffers from object duplication. For instance, it produces duplicated humans in rows 1 through 4 and a phantom figure in the grass background of row 5. By comparison, \gls{method} consistently generates high-quality images free from artifacts and duplication. This illustrates that \gls{method} effectively addresses a key limitation of prior approaches, enabling coherent and detailed high-resolution generation without duplication.

\subsection{Detail enhancement over the base image}
\figDetail
\autoref{fig:qualitative_comparison_sdxl} illustrates how \gls{method} enhances fine details and can improve semantic plausibility in base images generated by the SDXL model at the original resolution. Our method retains the same overall composition but successfully extracts and amplifies fine details that were merely suggested in the original SDXL generations. In row 1, the intricate blue patterns on the porcelain bottle show substantially improved definition, with clear brushwork details and subtle variations in the blue pigment that are barely distinguishable in the original image. Row 2 illustrates how our method reveals individual yarn strands and stitch patterns of a knitted toy mouse with great clarity. In row 3, a child in the flower field shows how our method can handle natural scenes with finer hair detail, clothing texture, and surrounding flora detail without losing scene composition. These results demonstrate that \gls{method} effectively adds the fine-grained details necessary for high-quality high-resolution image generation.

\subsection{Human evaluation study}
\figUserStduy
To validate our qualitative observations, we conducted a comprehensive human preference study comparing HiWave against Pixelsmith. Using 32 image pairs generated with identical prompts and seeds at 4096$\times$4096 resolution, we presented participants with randomized blind A/B tests. The 32 prompts were selected solely based on the SDXL base outputs--prior to running any methods--to avoid selection bias. The participants were asked to select their preferred image based on overall quality, coherence, and absence of artifacts.
\autoref{tab:barplot_preference} shows the preference scores for HiWave and Pixelsmith.
Across 548 independent evaluations, HiWave was preferred in 81.2\% of responses (445 out of 548), with seven test cases achieving 100\% preference for our method. The full set of per-question preference percentages is provided in \autoref{fig:full_human_eval} (Appendix).
% Pixelsmith was not preferred 100\% of the time in any case.
This strong preference aligns with our qualitative findings regarding artifact reduction and coherence preservation in \autoref{fig:qualitative_comparison}. 

\subsection{Ablation Study}
Additional results and ablation studies analyzing \gls{method}'s performance and the contribution of its components are provided in the appendix.
\autoref{fig:improvs_SDXL} further confirms that \ours improves both global structure and fine details compared to the base image generated with SDXL at 1024$\times$1024.
\autoref{fig:bands_configurations} demonstrates that removing guidance from low-frequency components effectively eliminates duplication artifacts.
\autoref{fig:noDDIM} highlights the importance of initializing sampling from DDIM-inverted noise to avoid geometric and color inconsistencies across patches,
while the bottom-right subfigure in \autoref{fig:bands_configurations} shows that this step alone is insufficient. Therefore, DWT-based frequency guidance remains essential for fully mitigating duplication and producing highly detailed, artifact-free images. % failedScores
\autoref{fig:oneShotvsMulti} confirms that multistep generation produces sharper details than single-step generation, without introducing duplication in our setup---unlike what has been reported in prior work \cite{tragakis2024one}.
\autoref{fig:realImagesUpscaled} illustrates that HiWave can also upscale natural (non-AI-generated) images in a zero-shot manner by leveraging the representations embedded in the SDXL noise space obtained via DDIM inversion.
\autoref{fig:8k} demonstrates HiWave’s ability to generate artifact-free images at an ultra-high resolution of 8192$\times$8192.
Lastly, additional comparisons in \autoref{fig:compPixelsmith} further shows that HiWave outperforms Pixelsmith by generating sharper, less blurry images while significantly reducing duplication artifacts.

\section{Conclusion}
In this work, we introduced HiWave, a novel zero-shot pipeline that enables pretrained diffusion models to generate ultra-high-resolution images (e.g., 4096×4096) beyond their native resolution, without requiring architectural modifications or additional training. HiWave employs a two-stage, patch-based strategy that first generates a base image using a pretrained diffusion model and subsequently refines individual patches of the upscaled image to achieve higher resolutions. Specifically, HiWave leverages a patch-wise DDIM inversion approach to recover the latent noise from the base image and incorporates a novel wavelet-based detail enhancement module that selectively guides high-frequency components while preserving or enhancing the global structure via low-frequency components. With this frequency-aware guidance mechanism, HiWave overcomes common pitfalls of existing high-resolution methods---namely, object duplication and structural incoherence---enabling models trained at 1024$\times$1024 to generate coherent and detailed outputs at 4096$\times$4096. Extensive experiments with Stable Diffusion XL demonstrated that HiWave not only produces visually compelling 4K images with fine details and strong global consistency, but also significantly outperforms prior zero-shot high-resolution approaches. A user study further supported these findings, with participants preferring HiWave's results in more than 80\% of cases. By enabling ultra-high-resolution synthesis without retraining, HiWave unlocks practical applications in domains where 4K (and beyond) outputs are essential. Future work could explore extensions to video, runtime optimizations, and the integration of more advanced guidance strategies within the detail enhancer module. Overall, HiWave represents a significant step toward democratizing high-fidelity, high-resolution generative modeling.

\bibliographystyle{ACM-Reference-Format}
\bibliography{hiwave-references}

\figMain
\figShowcase

\clearpage
\appendix

\section{Quantitative evaluation}\label{sec:quantitative-evaluation}

\begin{table*}[htp]
\caption{Quantitative comparison of high-resolution image generation methods. Results are based on 1,000 random samples from the LAION/LAION2B-en-aesthetic dataset. Although current metrics struggle to reliably evaluate high-resolution images, HiWave achieves comparable or superior performance to state-of-the-art methods in this domain.}
\begin{tabular}{clcccccc}
\toprule
\textbf{Resolution} & \textbf{Model} & \textbf{FID}$\downarrow$ & \textbf{KID}$\downarrow$ & \textbf{IS}$\uparrow$ & \textbf{CLIP}$\uparrow$ & \textbf{LPIPS}$\downarrow$ & 
\textbf{ HPS-v2 }$\uparrow$\\
\midrule
$1024^2$ & SDXL & \textbf{61.78}  & \textbf{0.0020} & 18.6651 & 33.2183 & \textbf{0.7779} & \textbf{0.2639} \\
\midrule
\multirow{3}{*}{$2048^2$} & HiWave (Ours) & 63.3460   & 0.0027 & \textbf{19.4784} & 33.2568 & 0.7784  & 0.2620 \\
 & Pixelsmith &  62.3054   & 0.002197 & 19.1222 & 33.1550 & 0.7808 & 0.2598 \\
  & HiDiffusion & 65.9073   & 0.002895 & 17.7217 & 31.9566 & 0.7834 & 0.2432 \\
  % & SDXL & 92.4854   & 0.010774 & 14.0658 & 31.3733 & 0.7828 & 0.2274\\
\midrule
\multirow{3}{*}{$4096^2$} & HiWave (Ours) & 64.7265    & 0.003241 & 18.7743 & \textbf{33.2729} & 0.7831 & 0.2585\\
% & HiWave+CC & 64.7538    & 0.003129 & 18.8701 & 33.2294 & 0.8139 &  0.2610\\
 & Pixelsmith & 62.5542  & 0.002373 & 19.4302 &  33.1504 & 0.8039 & 0.2598 \\
& HiDiffusion &  93.4473  &  0.014934  & 14.6960  & 28.2296  & 0.7996 & 0.1821 \\
\bottomrule
\end{tabular}
\label{tab:metrics}
\end{table*}

\autoref{tab:metrics} presents a quantitative comparison using several standard metrics commonly used in evaluating generative image models. While these metrics provide useful insights into model performance at moderate resolutions, it is important to recognize their limitations when assessing high-resolution outputs (e.g., 2048$\times$2048 and 4096$\times$4096) produced by methods such as HiWave.

A key issue is that many of these metrics depend on neural network architectures designed for fixed, lower-resolution inputs (See below for details). As a result, evaluating our high-resolution images requires substantial downsampling, which may obscure improvements in fine-grained details, textures, and overall fidelity that high-resolution synthesis is intended to achieve.

\begin{figure}[t]
  \centering
  \includegraphics[width=0.45\textwidth]{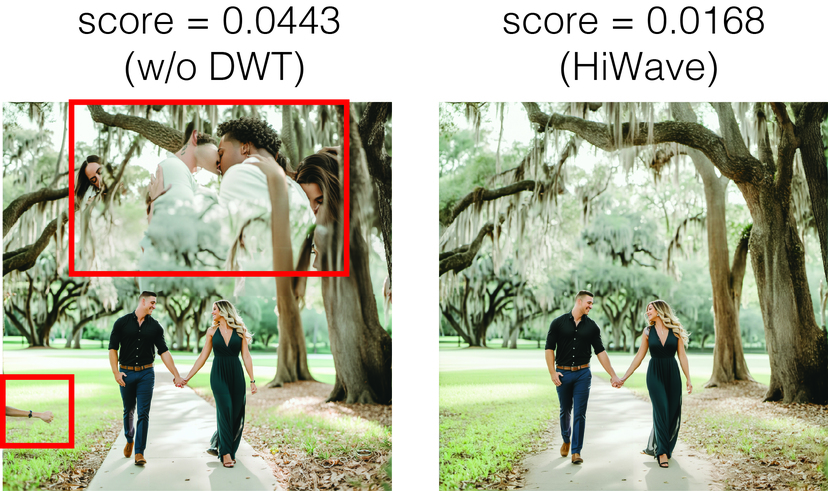}
  \caption{Failure example of current metrics in evaluating high-resolution images. The left image (HiWave w/o DWT) contains visible duplication artifacts, including a hallucinated couple in the trees and a disembodied hand in the grass. In contrast, the DWT-guided HiWave result (right) is clean and artifact-free. Despite this clear perceptual difference, common metrics faile to reflect the quality gap. ImageReward scores are 0.0443 vs 0.0168, respectively.}
  \label{fig:failedScores}
\end{figure}

Additionally, perceptual inconsistencies often go undetected by these scores. For instance, as shown in \autoref{fig:failedScores}, the version of the wedding photo generated without DWT guidance exhibits clear duplication artifacts---hallucinating a second couple in the tree canopy---whereas the DWT-guided version appears visually clean. However, metrics such as ImageReward (0.0443 vs 0.0168) fail to reflect this perceptual gap, further emphasizing the need for qualitative and human evaluation.

Despite these limitations, HiWave demonstrates comparable or superior performance relative to prior methods across both resolutions, underscoring its capacity to produce high-quality, detailed images.

We provide details of each metric used in \autoref{tab:metrics} below.

\textbf{FID (Fréchet Inception Distance)} \cite{heusel2017gans} and \textbf{KID (Kernel Inception Distance)} \cite{binkowski2018demystifying}: These widely used metrics compare feature representations extracted by the Inception V3 network \cite{szegedy2016rethinking}. Inception V3 requires input images to be resized to 299x299 pixels. Downsampling from 4096x4096 to 299x299 inevitably leads to a loss of high-frequency information.

\textbf{IS (Inception Score)} \cite{salimans2016improved}: Similarly, the Inception Score relies on the Inception V3 network and thus requires images to be resized to 299x299 pixels. While indicative of image quality at that scale, it does not assess details only visible at higher resolutions.

\textbf{CLIP Score}: This metric evaluates the semantic alignment between the generated image and the input text prompt using embeddings from the CLIP model \cite{radford2021learning}. Commonly used CLIP vision encoders, such as ViT-L/14 or ViT-B/32, process images at a resolution of 224x224 pixels. This significant downsampling primarily captures semantic content rather than high-resolution fidelity.

\textbf{LPIPS (Learned Perceptual Image Patch Similarity)} \cite{zhang2018unreasonable}: LPIPS measures perceptual similarity using features from networks like AlexNet or VGG. While designed to correlate well with human perception, standard implementations often process images resized to resolutions like 256x256 pixels, limiting their ability to assess fine details present only at much higher resolutions.

\textbf{HPS-v2 (Human Preference Score v2)} \cite{wu2023human}: While HPS v2 addresses limitations of earlier metrics by aligning better with human aesthetic judgment, it still relies on the CLIP ViT-H/14 architecture \cite{radford2021learning}, which processes images at 224x224 resolution. Therefore, evaluating our 4096x4096 outputs requires the same significant downsampling as the CLIP score, potentially losing some of the fine details our method aims to generate.

\section{Runtime analysis}

\begin{table}[t]
  \centering
  \caption{Inference time comparison (in seconds) for various high-resolution image generation methods on an RTX 3090 GPU. HiWave achieves a runtime comparable to previous patch-based generation approaches. Runtimes for prior methods are reported from \cite{tragakis2024one}. OOM indicates out-of-memory on the RTX 3090.}
  \maxsizebox{\columnwidth}{!}{
  \begin{booktabs}{colspec = {Q[l, m]Q[l, m]cc}}
    \toprule
     & \textbf{Method} & \textbf{2048²} & \textbf{4096²} \\
    \midrule
    \SetCell[r=4]{m} Direct inference & SDXL \Citep{sdxl} & 71 & 515 \\
     &  ScaleCrafter \cite{he2023scalecrafter} & 80 & 1257 \\
    & HiDiffusion \cite{zhang2023hidiffusion} & 50 & 255 \\
    & FouriScale \cite{huang2024fouriscale} & 162 & OOM \\
    \midrule
    \SetCell[r=3]{m} Patch-based & DemoFusion \cite{du2024demofusion} & 219 & 1632 \\
    & AccDiffusion \cite{lin2024accdiffusion} & 231 & 1710 \\
    & Pixelsmith \cite{tragakis2024one} & 130 & 549 \\
    & HiWave (Ours) & 238 & 1557 \\
    \bottomrule
  \end{booktabs}
  }
  \label{tab:inference_time}
  % \\ \small{*All measurements except for ours are from \cite{tragakis2024one}. Measured on NVIDIA RTX 3090, OOM: Out of memory on RTX 3090.}
\end{table}

\autoref{tab:inference_time} presents the inference times for various high-resolution generation methods. Our method delivers high-quality results with a reasonable runtime compared to previous approaches. While direct inference methods are generally faster, they often sacrifice visual quality at high resolutions. Patch-based methods, though slower, offer superior fidelity and detail. HiWave embraces this trade-off, prioritizing image quality over speed---an approach validated by strong qualitative outcomes. Although our runtime exceeds that of Pixelsmith, this is due to our multistep upscaling strategy designed to enhance fidelity (as illustrated in \autoref{fig:oneShotvsMulti}). In contrast, Pixelsmith’s one-shot pipeline prioritizes speed at the cost of quality. Considering these differences, our runtime remains comparable to other patch-based methods.

\section{Improvements over the base image}
\begin{figure}[t]
  \centering
  \includegraphics[width=0.9\linewidth]{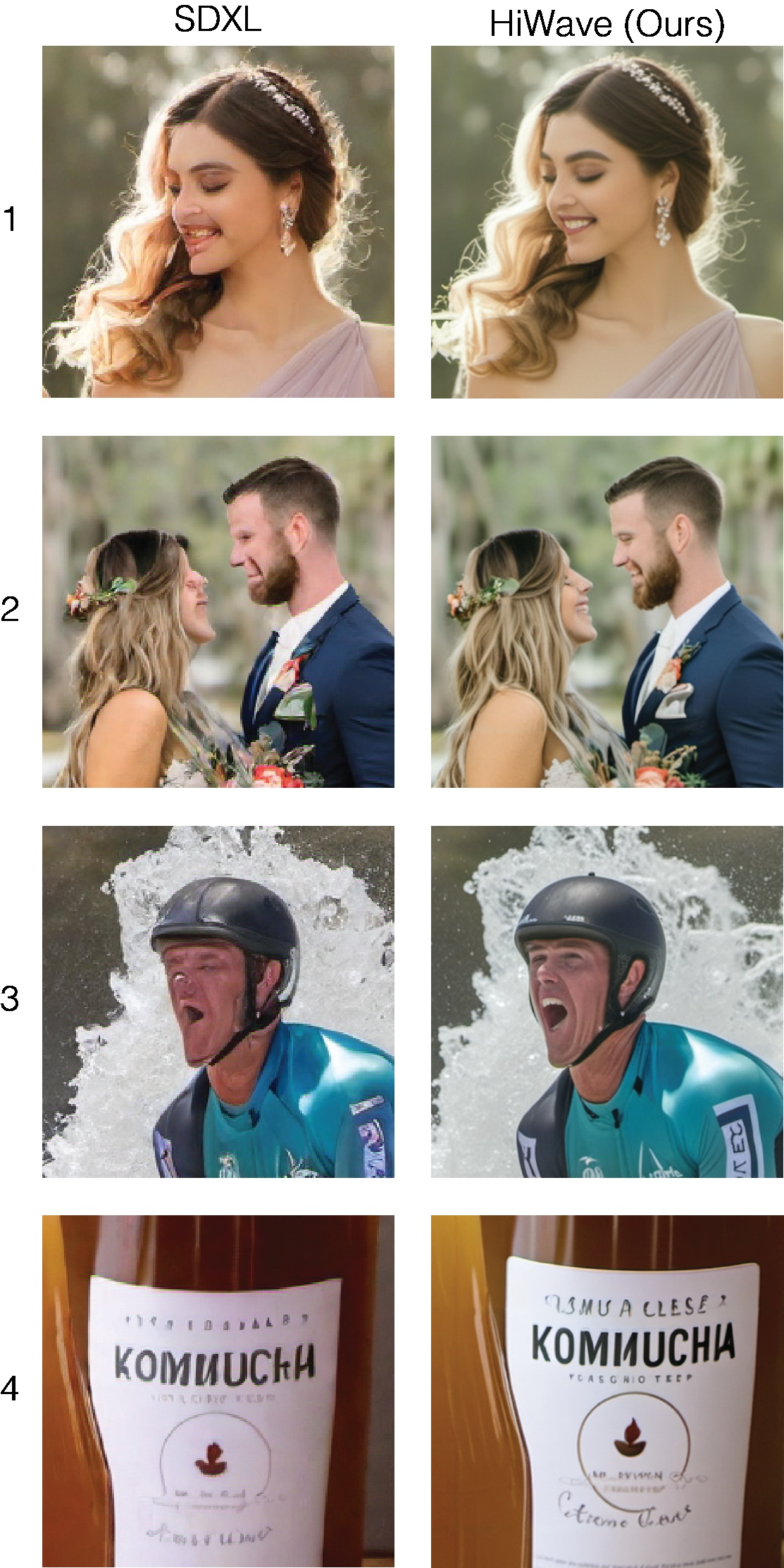}
  \caption{Comparison between SDXL base generations and HiWave generations across various challenging content types. The shown crops are 1/5 the length of the original images (1024$\times$1024 for SDXL, and  4096$\times$4096 for HiWave). Our approach enhances facial definition (Rows 1–3) and renders sharper characters (Row 4), while preserving the global composition and visual style.}
  \label{fig:improvs_SDXL}
\end{figure}

In \autoref{fig:improvs_SDXL}, we compare the base generations of SDXL with the HiWave outputs across a variety of challenging content types. The examples include cropped regions representing 1/5 the width of the original images (SDXL at 1024$\times$1024, HiWave at 4096$\times$4096). HiWave consistently enhances facial definition (Rows 1–3) and produces sharper character renderings (Row 4), while faithfully preserving the global composition and visual style of the original outputs. These results demonstrate that HiWave effectively improves both the fine-grained details and structural coherence of the base images.

\subsection{Analyzing the role of frequency-specific guidance}
\begin{figure}[t]
  \centering
  \includegraphics[width=0.9\linewidth]{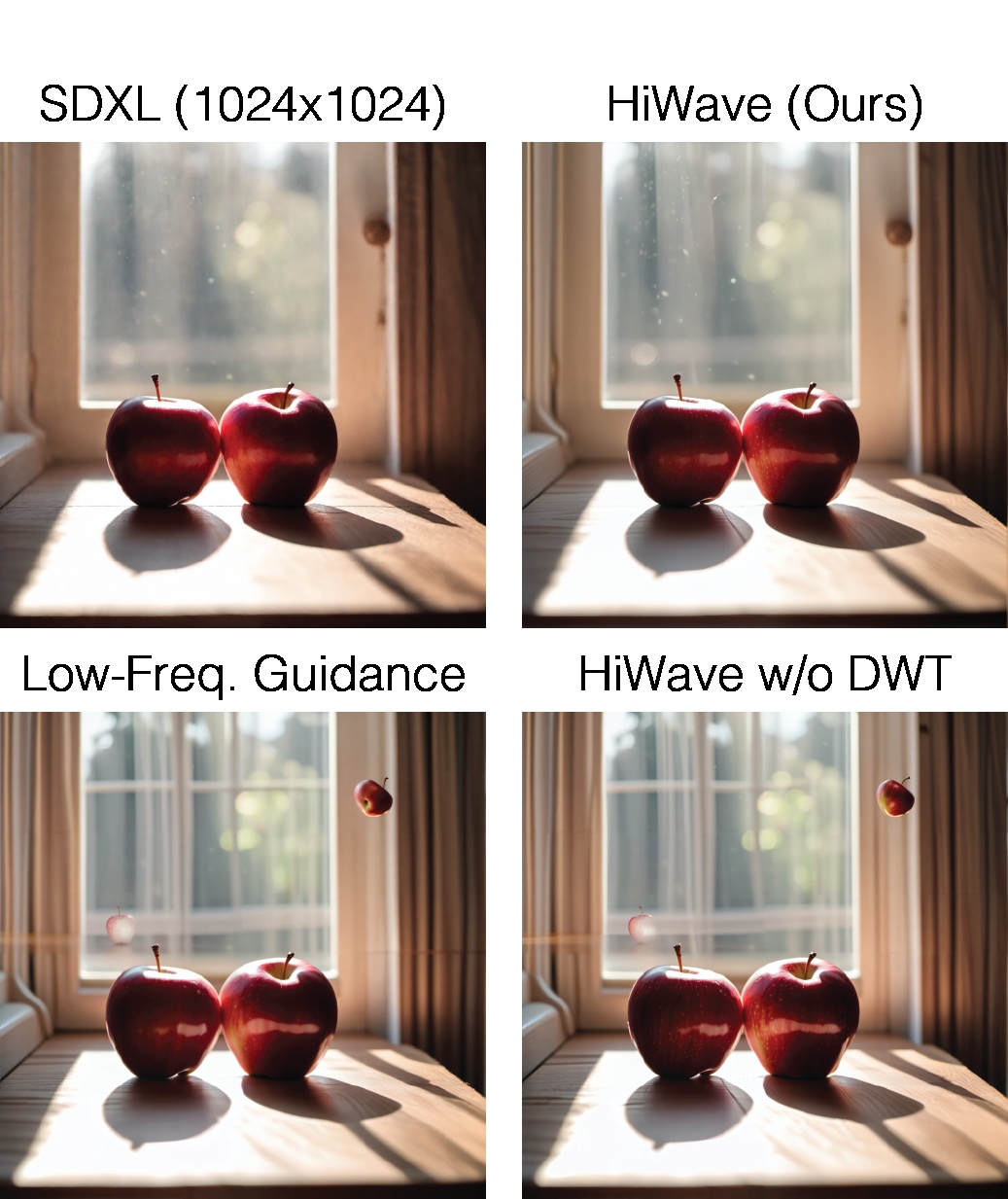}
  \caption{Comparison of different configurations for our DWT-based frequency guidance. Top-left: SDXL base image at 1024×1024. Top-right: full HiWave with DWT-based guidance at 4096×4096. Bottom-left: guidance applied only to low-frequency bands (inverse of our method). Bottom-right: standard CFG. Our DWT-based approach is crucial for reducing duplications and enhancing details.}
  \label{fig:bands_configurations}
\end{figure}

\autoref{fig:bands_configurations} compares various configurations of our frequency guidance strategy. The top-left image shows the baseline SDXL output at 1024$\times$1024 resolution, while the top-right image presents the result of our full HiWave method using DWT-based guidance at 4096$\times$4096. To isolate the effect of frequency-specific control, the bottom-left variant applies guidance only to low-frequency bands---essentially the inverse of HiWave---while the bottom-right replaces our DWT-based approach with standard CFG. Both the low-frequency-only and CFG-only variants exhibit noticeable duplication artifacts. In contrast, HiWave maintains the global composition of the original SDXL image while enhancing fine-grained details. These results suggest that maintaining low-frequency components while selectively guiding high frequencies is crucial for producing coherent and detailed generations.

\section{Importance of DDIM inversion}
\begin{figure}[t]
  \centering
  \includegraphics[width=0.9\linewidth]{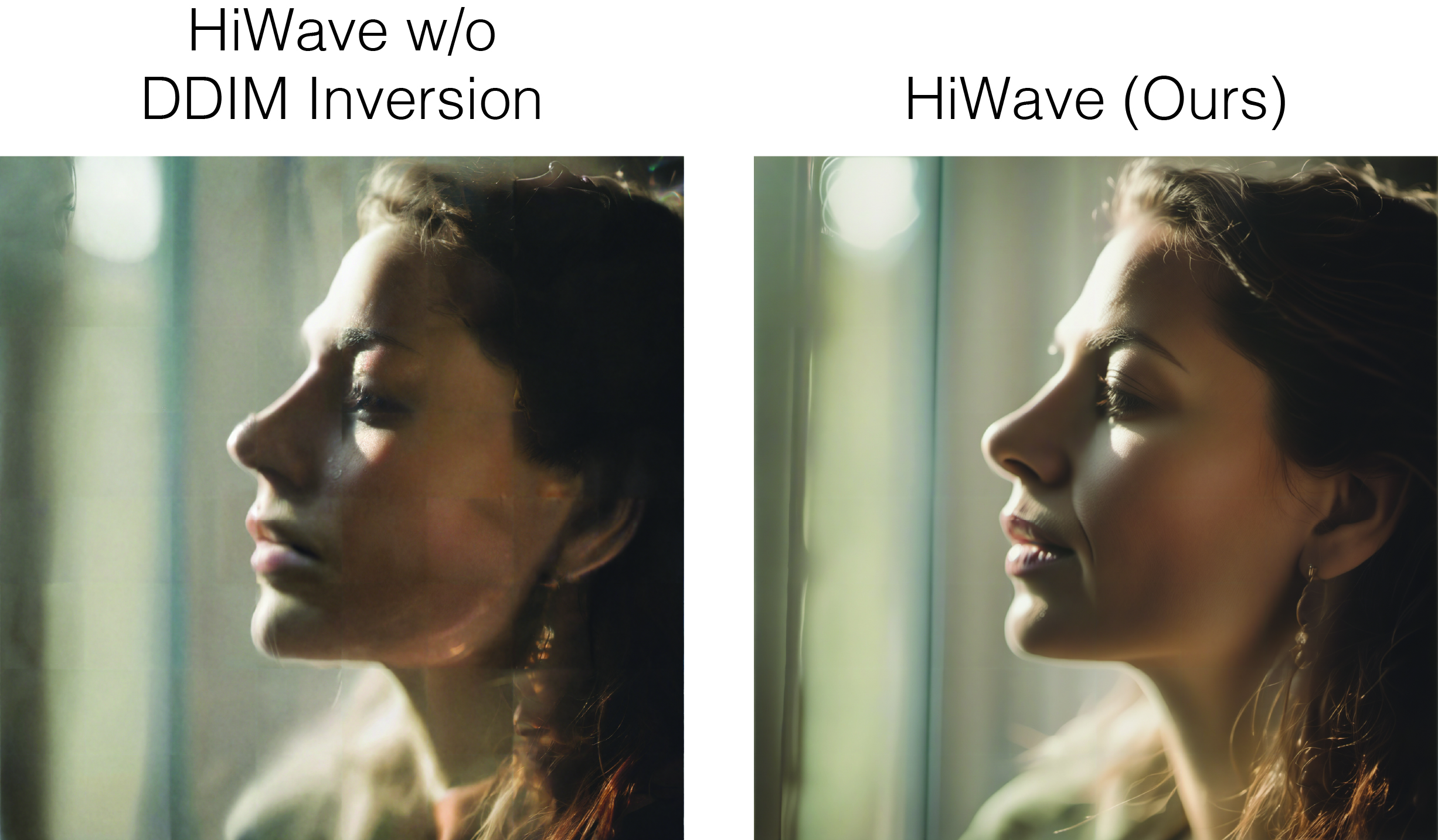}
  \caption{Effect of omitting DDIM inversion in HiWave. Left: HiWave without DDIM inversion. Right: Full HiWave method. Without DDIM inversion, patch-wise generation becomes independent, resulting in visible seams, inconsistent geometry, and color mismatches across patches.}
  \label{fig:noDDIM}
\end{figure}

To illustrate the impact of DDIM inversion in HiWave, \autoref{fig:noDDIM} compares outputs with and without this component. The left image shows HiWave applied without DDIM inversion, while the right displays the result of the full HiWave method. Without DDIM inversion, patch-wise generation becomes uncoordinated and independent, resulting in visible seams, geometric inconsistencies, and color mismatches at patch boundaries. In contrast, the full HiWave method preserves spatial coherence across the entire image.

\section{One-Shot vs multistep generation}

\begin{figure}[t]
  \centering
  \includegraphics[width=0.9\linewidth]{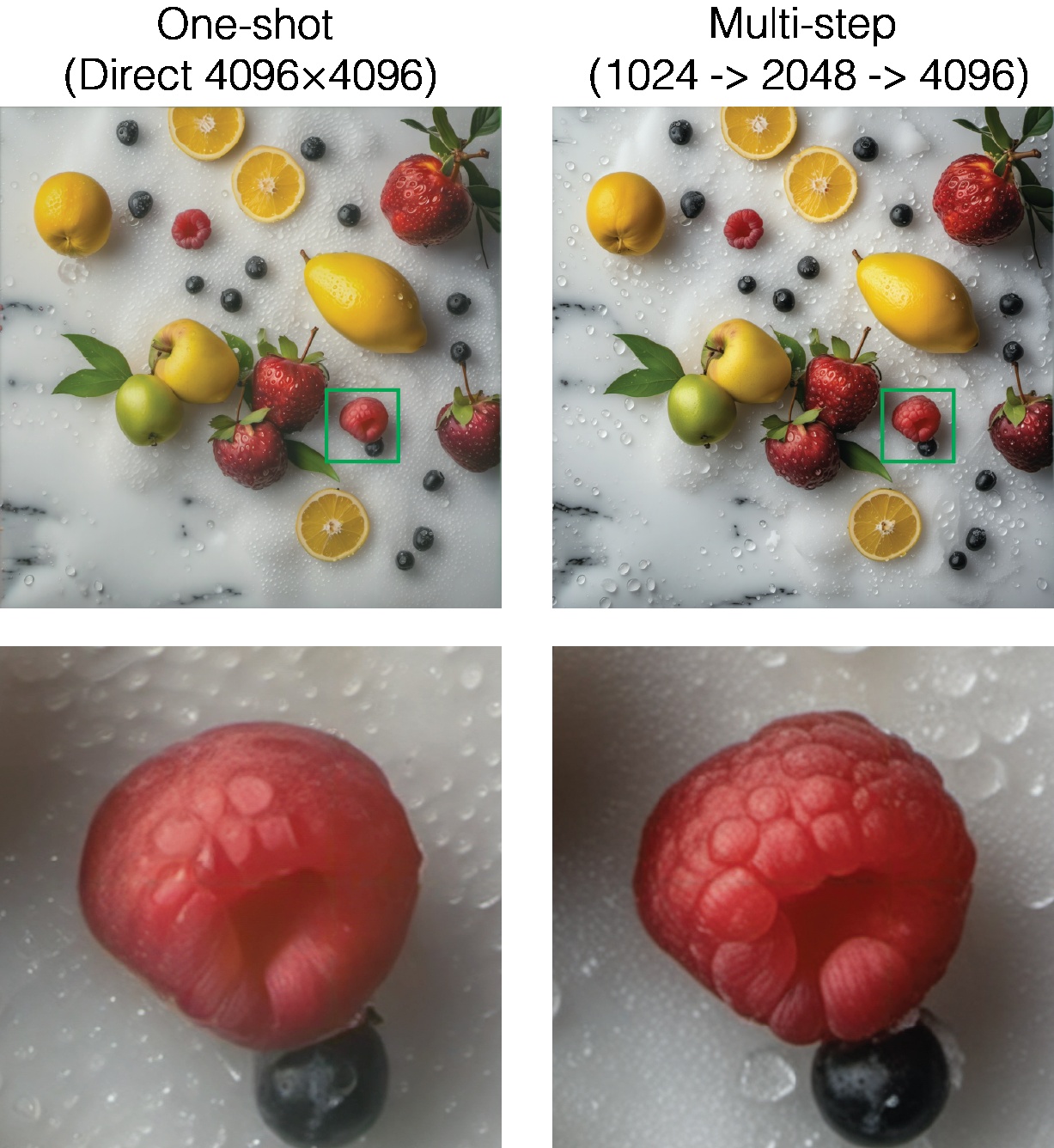}
  \caption{Comparison between single-step 4096×4096 generation (left) and progressive multistep upscaling (right). The multistep approach generates the image at 1024, then progressively upscales it to 2048 and 4096. This progressive generation process results in sharper details and enhanced overall quality.}
  \label{fig:oneShotvsMulti}
\end{figure}
This experiment compares one-shot generation at 4096$\times$4096 resolution with a progressive multistep upscaling strategy using HiWave. In the one-shot setting, the image is generated directly at the target resolution. In the multistep approach, generation begins at 1024$\times$1024 and is successively upscaled to 2048$\times$2048 and then to 4096$\times$4096. As illustrated in \autoref{fig:oneShotvsMulti}, the multistep strategy yields higher fidelity. For example, the raspberry exhibits a more realistic surface and texture, whereas the one-shot result lacks fine details. These findings highlight the importance of progressive upscaling for high-resolution synthesis, as each intermediate step allows HiWave to incrementally refine and enrich image details.

\section{Upscaling real (non-generated) images}
\figUpscaleReal
% \documentclass{article}
% \usepackage{pgfplots}
% \pgfplotsset{compat=1.18}
% \usepackage{filecontents}

% CSV data as inline file

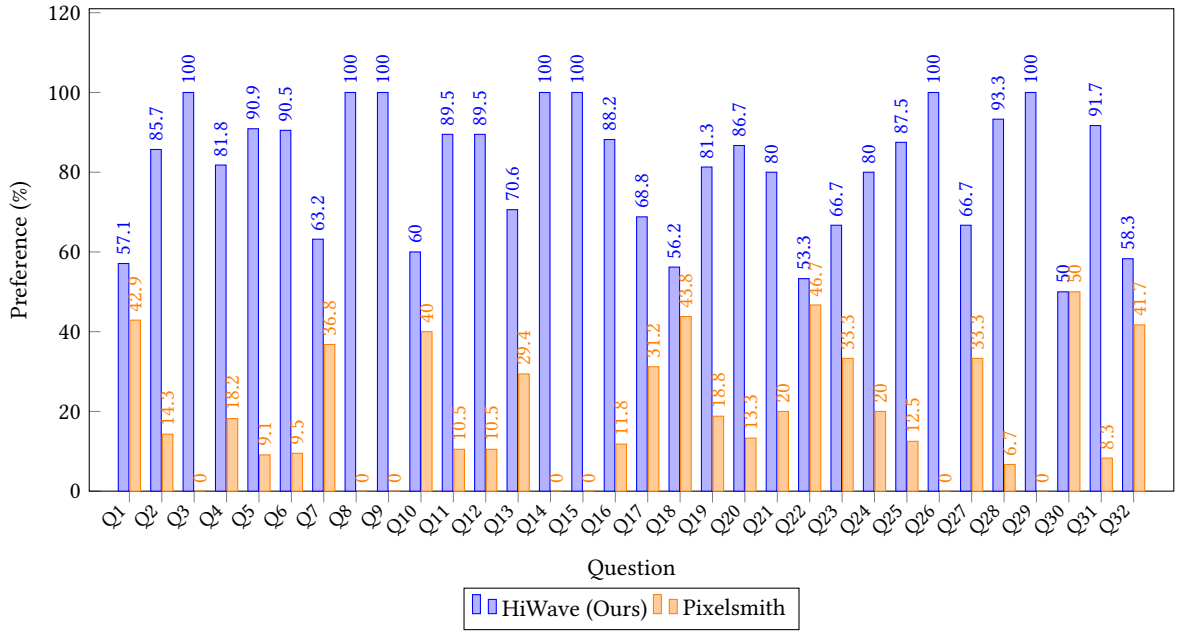
\begin{figure*}[htbp]
\centering
\begin{tikzpicture}
\begin{axis}[
    ybar,
    bar width=4pt,
    width=16cm,
    height=8cm,
    enlarge x limits=0.04,
    enlarge y limits={upper,value=0.1},
    ylabel={Preference (\%)},
    xlabel={Question},
    ymin=0, ymax=110,
    legend style={at={(0.5,-0.2)}, anchor=north, legend columns=-1},
    symbolic x coords={Q1,Q2,Q3,Q4,Q5,Q6,Q7,Q8,Q9,Q10,Q11,Q12,Q13,Q14,Q15,Q16,Q17,Q18,Q19,Q20,Q21,Q22,Q23,Q24,Q25,Q26,Q27,Q28,Q29,Q30,Q31,Q32},
    xtick=data,
    xticklabel style={rotate=45, anchor=east, font=\small},
    nodes near coords,
    xtick pos=left,
    ytick pos=left,
    % nodes near coords align={vertical},
    nodes near coords style={
        font=\small,
        rotate=90,
        anchor=west,
        /pgf/number format/fixed,
        /pgf/number format/precision=1
    },
]
\addplot+[style={blue,fill=blue!30}, bar shift=-2.2pt] table[x=Q,y=Ours,col sep=comma]{preference_data.csv};
\addplot+[style={orange,fill=orange!40}, bar shift=2.2pt] table[x=Q,y=Pixelsmith,col sep=comma]{preference_data.csv};
\legend{\ours (Ours), Pixelsmith}
\end{axis}
\end{tikzpicture}
\caption{Human preference (\%) across 32 image pairs comparing HiWave and Pixelsmith. Blue bars indicate the proportion of votes favoring HiWave, while orange bars represent votes for Pixelsmith in the blind A/B test. HiWave significantly outperforms Pixelsmith in most cases, even achieving 100\% preference in 7 instances.}
\label{fig:full_human_eval}
\end{figure*}

To evaluate the applicability of HiWave beyond synthetic images, we applied our method to real 1024$\times$1024 photographs instead of generated images from SDXL. For these real photographs, we manually crafted descriptive prompts that matched the image content, as such prompt conditioning is necessary for HiWave to guide the high-resolution generation. \autoref{fig:realImagesUpscaled} compares the original photographs (left) with their corresponding high-resolution upscalings to 2048$\times$2048 using HiWave. In the turtle example, HiWave convincingly enhances fine details such as the shell texture and flipper structure. The improvement is subtler for the market scene, possibly due to domain-specific differences in style or tone compared to the SDXL training distribution. In the final example, HiWave enhances structural elements like the hand and phone and sharpens some text regions, though it may also introduce minor variations in fine details—an area that could benefit from further refinement. Overall, these results demonstrate HiWave's strong potential for real-image enhancement, even in challenging scenarios.

\section{Detailed results of our user study}
To evaluate perceptual quality, we conducted a blind A/B test comparing HiWave to Pixelsmith across 32 diverse image pairs. Participants were asked to indicate their preferred image in each pair. As shown in \autoref{fig:full_human_eval}, HiWave was preferred in all cases and received a substantially higher proportion of votes in the majority of comparisons. This strong preference shows the effectiveness of our approach in producing visually compelling outputs and demonstrates consistent perceptual superiority over previous methods.

\section{Scaling HiWave to 8K resolution}
\begin{figure*}[tp]
  \centering
  \includegraphics[width=0.95\textwidth]{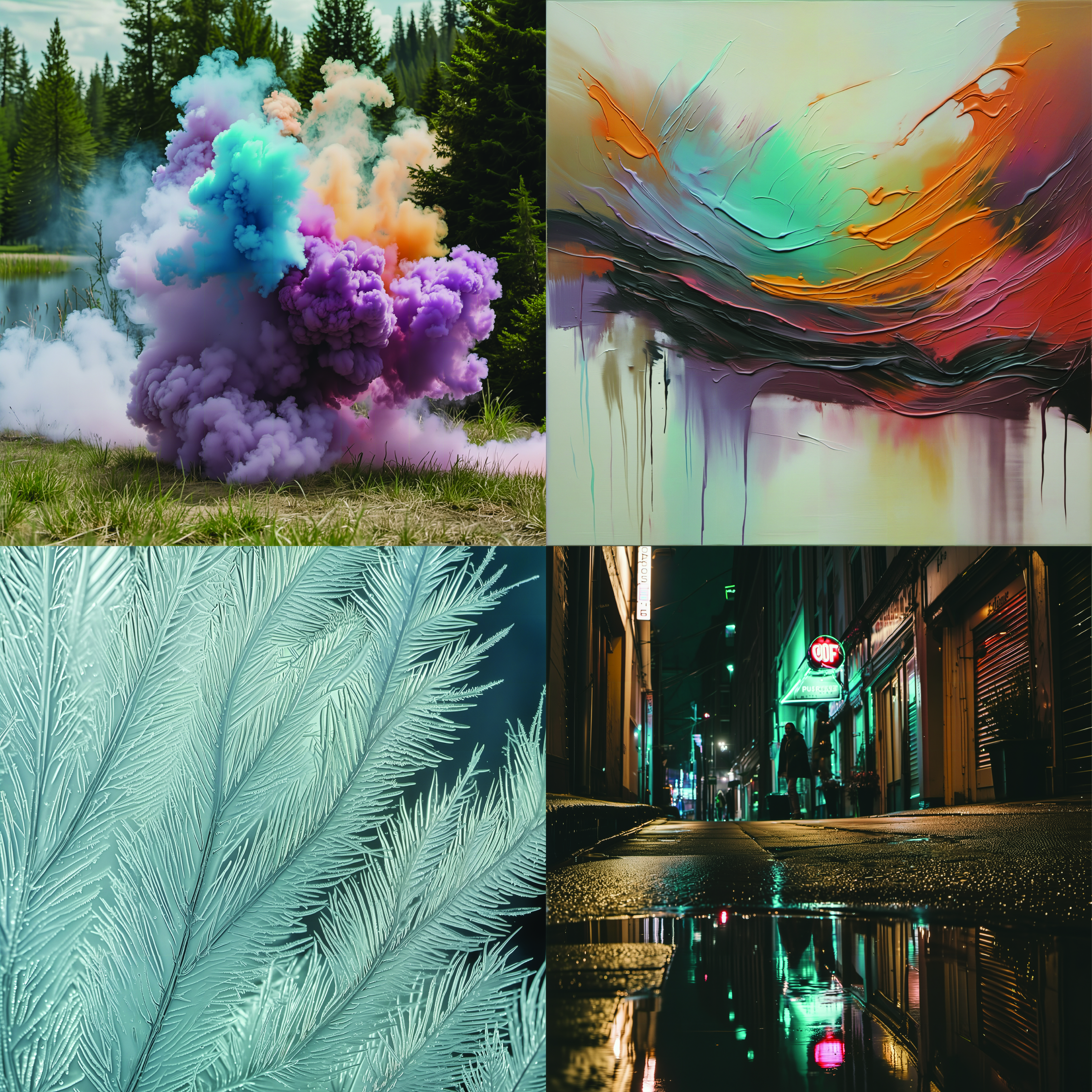}
  \caption{8K generation results using HiWave. The model continues to produce plausible images with fine details and coherent structure at this ultra-high resolution. This experiment demonstrates HiWave’s ability to scale effectively while preserving structural coherence.}
  \label{fig:8k}
\end{figure*}

To examine the scalability limits of HiWave, we extended our method to generate images at 8192$\times$8192 resolution---64 times the pixel count of the standard 1024$\times$1024 baseline. As shown in \autoref{fig:8k}, HiWave successfully maintains structural coherence and fine-grained details even at this extreme resolution. Its ability to preserve consistency and quality at such scale highlights the robustness of our approach, positioning it as a promising tool for applications demanding ultra-high resolutions.

\section{More Qualitative comparisons with Pixelsmith}
\begin{figure*}[tp]
  \centering
  \includegraphics[width=0.92\textwidth]{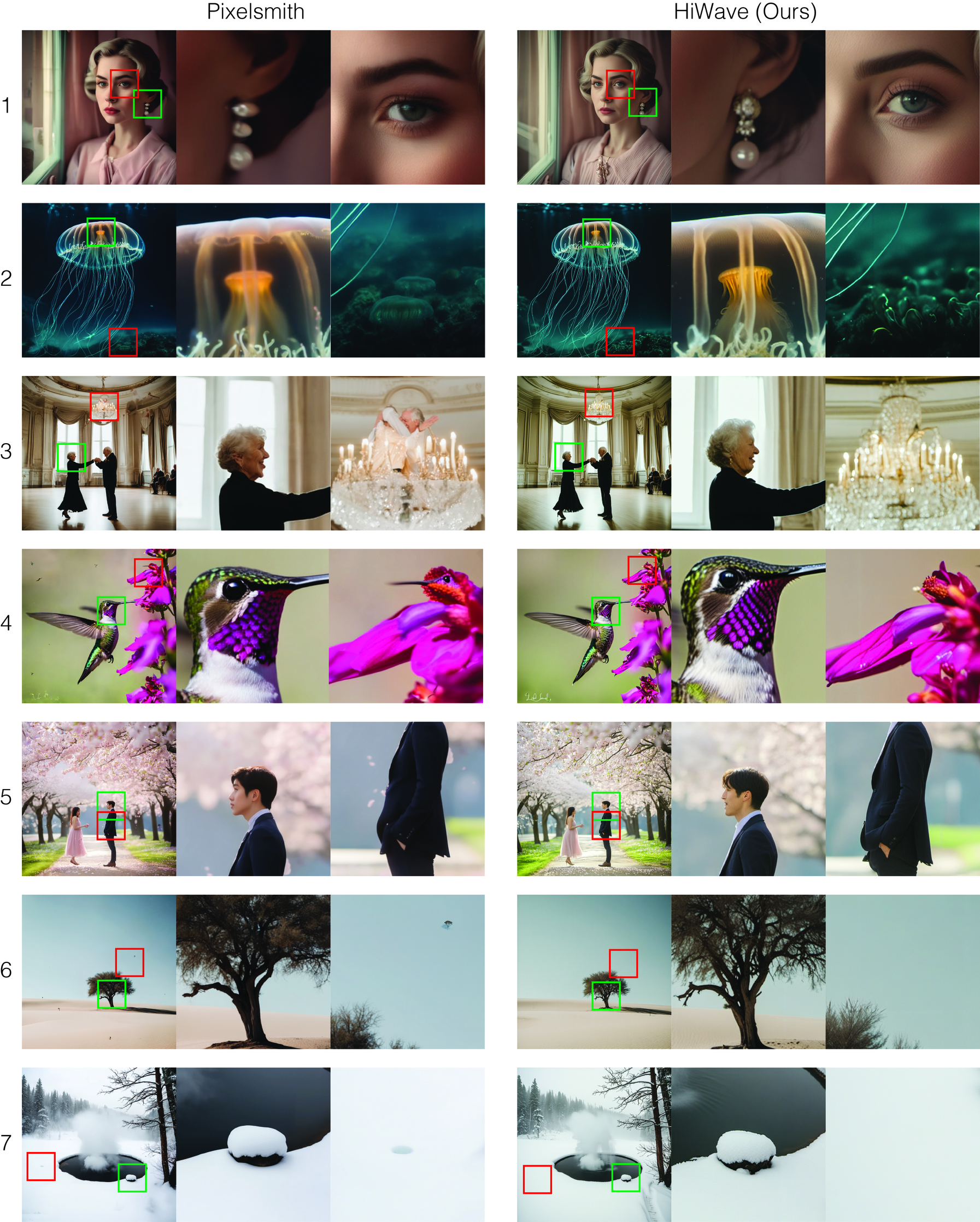}
  \caption{Furthe comparison of HiWave with Pixelsmith across diverse 4096$\times$4096 scenes. Each row presents a different scene, with corresponding crops for detailed inspection. HiWave consistently preserves global structure while enhancing local textures and edge sharpness. In several cases, HiWave reduces blurring, improves fine details, and avoids duplications seen in Pixelsmith outputs.}
  \label{fig:compPixelsmith}
\end{figure*}

In this section, we present additional qualitative comparisons between HiWave and Pixelsmith on 4096$\times$4096 resolution scenes to further evaluate the visual quality of our method. As shown in \autoref{fig:compPixelsmith}, each row displays a different scene along with close-up crops for detailed inspection. HiWave consistently preserves global structural integrity while enhancing local textures and edge sharpness. Compared to Pixelsmith, our method more effectively reduces blurring, retains fine details, and avoids common artifacts such as duplications. These results further highlight HiWave’s advantage in generating high-fidelity images.

% \section{Effect of Applying DWT on Noise vs\ $x_0$ Predictions}

% \begin{figure}[htp]
%   \centering
%   \includegraphics[width=0.95\linewidth]{assets/Appendix/x0_vs_noise}
%   \caption{The left column applies our DWT guidance to the predicted $x_0$ image before converting it back to noise, while the right column applies our DWT guidance directly to the noise prediction. While both approaches yield similar overall details, guiding the noise prediction helps reduce duplications better. For example, in the $x_0$ guided version (left), an additional faint human figure appears, whereas the Noise guided result (right) avoids this and produces a cleaner, more coherent structure in the ferns.}
%   \label{fig:noise_dwt}
% \end{figure}

\clearpage

\end{document}